\documentclass[twoside,11pt]{article}
\usepackage[abbrvbib, preprint]{jmlr2e}

\usepackage{amsmath,amsfonts,bm}

\def\eqref#1{equation~\ref{#1}}

\def\1{\bm{1}}

\DeclareMathAlphabet{\mathsfit}{\encodingdefault}{\sfdefault}{m}{sl}
\SetMathAlphabet{\mathsfit}{bold}{\encodingdefault}{\sfdefault}{bx}{n}

\usepackage{hyperref}
\usepackage{url}

\usepackage[utf8]{inputenc} \usepackage[T1]{fontenc}    \usepackage{hyperref}       \usepackage{url}            \usepackage{booktabs}       \usepackage{amsfonts}       \usepackage{nicefrac}       \usepackage{microtype}

\usepackage[export]{adjustbox}

\usepackage{amsmath}       \usepackage{graphicx}      
\usepackage{wrapfig}      
\usepackage{soul}      

\usepackage{algorithm}
\usepackage{algorithmic}
\usepackage{optidef}
\usepackage{tabularx}
\usepackage{multirow}
\usepackage[export]{adjustbox}
\usepackage{subfig}
\usepackage{mathtools}

\usepackage{bbm}

\newcommand{\MNNs}{Mechanistic Neural Networks}
\newcommand{\mnnblock}{mechanistic block}
\newcommand{\MnnBlock}{Mechanistic Block}
\newcommand{\MnnBlocks}{Mechanistic Blocks}
\newcommand{\Mnnblocks}{Mechanistic blocks}
\newcommand{\mnnblocks}{mechanistic blocks}
\newcommand{\mnnencoder}{mechanistic encoder}

\newcommand{\mnnrep}{ODE representation}
\newcommand{\mnnreps}{ODE representations}

\ShortHeadings{Mechanistic Neural Networks}{Mechanistic Neural Networks}
\firstpageno{1}

\begin{document}

\title{Mechanistic Neural Networks for Scientific Machine Learning}

\author{\name Adeel Pervez \email a.a.pervez@uva.nl \\
       \addr Informatics Institute,\\
       University of Amsterdam\\
       Amsterdam, The Netherlands
       \AND
       \name Francesco Locatello \email Francesco.Locatello@ist.ac.at\\
       \addr Institute of Science and Technology\\
       Klosterneuburg, Austria 
       \AND
       \name Efstratios Gavves \email e.gavves@uva.nl \\
       \addr Informatics Institute,\\
       University of Amsterdam\\
       Amsterdam, The Netherlands }

\maketitle
\begin{abstract}

\looseness=-1This paper presents \emph{\MNNs} -- a neural network design for machine learning applications in the sciences. It incorporates a new \textit{\MnnBlock} in standard architectures to explicitly learn governing differential equations as representations, revealing the underlying dynamics of data and enhancing interpretability and efficiency in data modeling.
Central to our approach is a novel \emph{Relaxed Linear Programming Solver} (NeuRLP) inspired by a technique that reduces solving linear ODEs to solving linear programs. This integrates well with neural networks and surpasses the limitations of traditional ODE solvers enabling scalable GPU parallel processing.
Overall, \MNNs\space demonstrate their versatility for scientific machine learning applications, adeptly managing tasks from equation discovery to dynamic systems modeling. We prove their comprehensive capabilities in analyzing and interpreting complex scientific data across various applications, showing significant performance against specialized state-of-the-art methods.
\footnote{Source code is available at \url{https://github.com/alpz/mech-nn}}

 \end{abstract}

\section{Introduction}
\label{sec:intro}

\looseness=-1Understanding and modeling the mechanisms underlying the evolution of data is a fundamental scientific challenge and is still largely performed by hand by domain experts, who leverage their understanding of natural phenomena to obtain equations. 
This process can be time-consuming, error-prone, and limited by prior knowledge. 
In this paper, we introduce \emph{\MNNs}, a new neural network design that contains one or more \emph{\MnnBlock} that explicitly integrate governing equations as symbolic elements in the form of \mnnreps. To efficiently train them, we revisit classical results on linear programs~\citep{young1961linear,rabinowitz_applications_1968} and develop a GPU-friendly solver. Together, they enable automating the discovery of best-fitting mechanisms from data in an efficient, scalable, and interpretable way.

\begin{figure}[t]
\centering
    \resizebox{\linewidth}{!}{
\label{tab:learning-dynamics}
\begin{tabular}{lcccc}
\toprule
          & Neural ODE,UDE  & SINDy  & Neural Operators  & Mech. NN \\
& \citet{chen2018neural} & \citet{brunton2016discovering} & \citet{li2020fourier} \\
& \citet{rackauckas2020universal} & & \\
\midrule
{Linear discovery}        & -- & $\checkmark$ & -- & $\checkmark$ \\
{Nonlinear discovery} & -- & -- & -- & $\checkmark$ \\
Physical parameters & $\checkmark$ & $\checkmark$ & -- & $\checkmark$ \\
Forecasting & $\checkmark$ & -- & $\checkmark$ & $\checkmark$ \\
Interpretability   & -- & $\checkmark$ & -- & $\checkmark$ \\ 
\bottomrule
\end{tabular}}
\vskip 0.05in
     \includegraphics[width=0.9\linewidth]{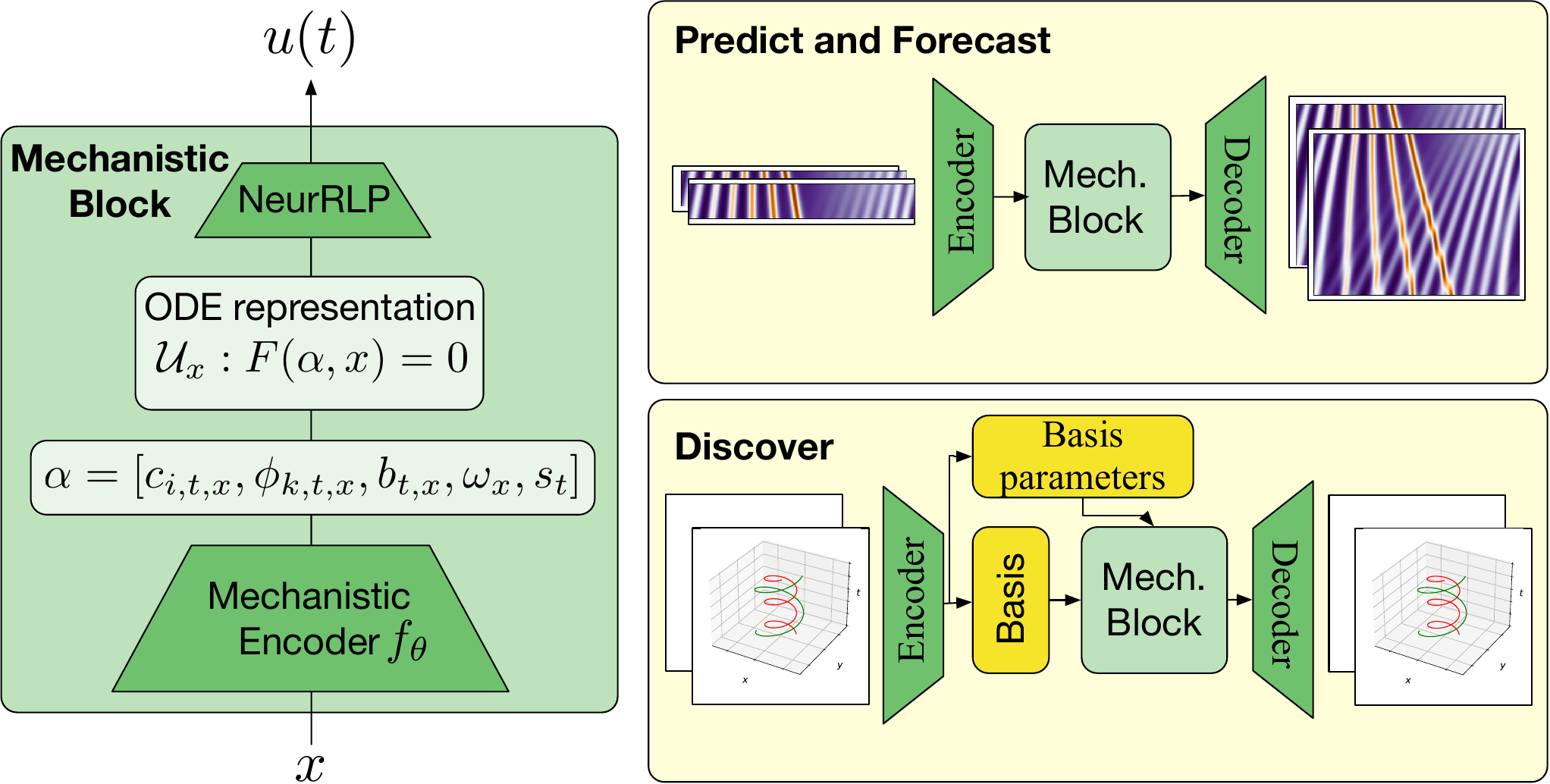}
\caption{\MNNs\space are a new neural network design that learn explicit \mnnreps. \MnnBlocks\space can used as bottlenecks in other neural networks to approximate dynamical systems and discover governing equations underlying data. Additional encoders and decoders are optional and depend on the application.}
    \label{fig:main}
\end{figure}

\looseness=-1 \MNNs{} present a fundamentally different computing paradigm than standard neural networks that rely on scalar or vector-valued numerical representations as their building block. They are composed of two parts: a \emph{\mnnencoder}{} and a \emph{solver}.
The output of the \mnnencoder{} is an explicit symbolic ``\textit{\mnnrep}''{} $\mathcal{U}_x$ of the general form
\begin{align}
\alpha &= f_\theta(x) \label{eq:mnn-1} \\ 
\mathcal{U}_x &= F(\alpha, x). \label{eq:mnn-2}
\end{align}

In more detail, $\mathcal{U}_x$ is a family of ordinary differential equations $F(\alpha, x)=0$, governed by learnable coefficients $\alpha$ that can be time-dependent or time-independent.
Coefficients $\alpha$ are obtained from the \mnnencoder~ $f_\theta$, and parameters $\theta$ are trained to optimally model the evolution of data $x=[x_1, ..., x_t]$ over time.
Unlike symbolic regression methods like SINDy~\citep{brunton2016discovering}, \Mnnblocks~can be stacked hierarchically in neural networks, and thus, the first challenge is that we must be able to train parameters and return accurate coefficients $\alpha$ for the hidden \mnnreps.
As direct supervision on the coefficients $\alpha$ is not available,
mechanistic layers are followed by differentiable solvers. This is the second challenge, as autoregressive ODE solvers are inefficient and will exhibit noisier gradients due to accumulating errors over long rollouts. When training \mnnreps, we must simultaneously learn the precise form of multiple independent ODEs (or independent systems of ODEs) and solve them over several time steps.
Sequential numerical solvers such as Runge-Kutta used in Neural ODEs~\citep{chen2018neural} are simply too inefficient for solving large batches of independent ODEs, as required for Mechanistic Neural Networks.

With \MNNs, we address both challenges directly in a ``native'' neural network context, as shown in Figure~\ref{fig:main}.
Building on an early method that reduces linear ODEs to linear programs~\citep{young1961linear}; combined with progress on differentiable optimization~\citep{amos2017optnet, wilder_melding_2018}; and also combined with recent developments that integrate fast parallel solutions of large constrained optimization problems in neural layers~\citep{pervez2023differentiable}, we propose a novel and ``natively'' neural \emph{Relaxed Linear Programming Solver} (NeuRLP) for ODEs.
NeuRLP has three critical advantages over traditional solvers, leading to more efficient learning over longer sequences than traditional sequential solvers.
These are: (i) \emph{step parallelism}, i.e., being able to solve for hundreds of ODE time steps in parallel, allowing for faster solving and efficient gradient flow; 
(ii) \emph{batch parallelism}, where we can solve in parallel on GPU batches of independent systems of ODEs in a single forward pass; (iii) \emph{learned step sizes}, where the step sizes are differentiable and learnable by a neural network.
The NeurRLP solver is parallel, scalable, and differentiable and can be extended with nonlinear dynamical loss terms for nonlinear ODEs.
Thus, the NeurRLP solver is ideal for training efficiently with neural networks that model complex ODEs, be it in their input and output or their intermediate hidden neural representations.

\paragraph{Relevance for scientific applications.} Machine Learning for dynamical systems has adopted specialized methodologies. Physics-Informed Neural Networks \citep{raissi2019physics} can be used for solving PDEs, data-driven neural operators \citep{li2020fourier} for forecasting and PDEs, linear regression on polynomial basis functions for discovering governing equations~\citep{brunton2016discovering, rudy2017data}, Neural ODEs~\citep{chen2018neural} for interpolation and control of dynamical systems \citep{doi:10.1137/21M1411433}. 
Being able to weave in governing equations in neural representations and solve them efficiently, \MNNs~ potentially offer a stepping stone for broad scientific applications of machine learning for dynamical systems (see table from Figure~\ref{fig:main}).
To empirically validate the claim and showcase the power, versatility, and generality of \MNNs, we perform a large array of experiments comparing and consistently outperforming the specialized golden standards: SINDy for equation discovery \citep{brunton2016discovering} (Section \ref{sec:mnn-discovery}), Neural ODE variants \citep{chen2018neural, norcliffe2020second} for modelling linear and nonlinear dynamics (Sections \ref{sec:mnn-nbody}, \ref{sec:mnn-time-series}, \ref{sec:mnn-discovery-physical}), and Neural operators \citep{li2020fourier, brandstetter2022lie}  for PDE modelling (Section \ref{mnn:pde-modeling}). 

\looseness=-1

\section{Mechanistic Neural Networks}
\label{sec:mnn}

We first describe the model for a \MnnBlock~ and leave the description of the \emph{Neural Relaxed Linear Program} solver for Section~\ref{sec:method}.

Formally, a \mnnencoder~in a \mnnblock~takes an input $x$ and generates a differential equation $\mathcal{U}_x$ as representation according to equations~\ref{eq:mnn-1} and~\ref{eq:mnn-2}.
The family of ordinary differential equations $\mathcal{U}_x : F(\alpha, x)=0$
\vskip -0.5cm
\begin{small}
\begin{align}
\overbrace{\sum_{i=0}^d c_{i}(t;x) u^{(i)}}^{\mathclap{\text{Linear Terms}}} + \overbrace{\sum_{k=0}^r \phi_k(t;x) g_k(t, u,u',\ldots )}^{\mathclap{\text{Non-Linear Terms}}} = b(t;x), \label{eq:ode-general}
\end{align}
\end{small}represents a broad parameterization for the \mnnrep, with an arbitrary number $d$ of linear terms with derivatives $u^{(i)}$ and an arbitrary number $r$ of nonlinear terms $g_k$ including derivatives $u^{(k)}, k=1, ..., d$.
We drop the obvious dependency of $u, u', ...$ on time $t$ to reduce notation clutter.
The coefficients $c_i(t;x), \phi_k(t;x)$ for the linear and nonlinear terms are functions that possibly depend on the time variable $t$ (thus non-autonomous ODEs), and on input $x$.
Furthermore,  $u$ may also be multidimensional, in which case equation \ref{eq:ode-general} would be a system of ODEs.
For clarity, we assume in the description a single dimension for the ODEs.

After computing the \mnnrep~ $\mathcal{U}_x$, we solve it with our specially designed parallel solver \emph{NeuRLP} for $n$ time steps and get a numerical solution as output of the \mnnblock: $z = \text{\texttt{solve\_ode}}(\mathcal{U}_x, \omega_x, n)$. The $\omega_x$ includes initial or boundary conditions and variables controlling step sizes that can also be specific to the input $x$. $\omega_x$ can be learned by NeurRLP, unlike traditional solvers.

\paragraph{\MnnBlocks\space in discrete time. \ }
Equations~\ref{eq:mnn-1}--~\ref{eq:ode-general} provide the mathematical description of \mnnrep~in \mnnblocks~in continuous time.
To implement them in a neural network, which is by nature discrete, we  discretize the continuous coefficients, parameters, function values, and derivatives in the ODE of~\ref{eq:ode-general},
\begin{align}
\sum_{i=0}^d c_{i, t, x} u^{(i)}_t + \sum_{k=0}^r  \phi_{k,t,x} g_k(t, u_t,u_t',\ldots)) &= b_{t, x} \label{eq:ode-general-discrete} \\
\text{s.t. \;}  [u_{t=1}, u_{t=1}', ...]&=\omega_x \label{eq:ode-general-discrete-constraints}
\end{align}
at discrete times $t=1, ..., n$ and with $n-1$ time steps $s_t$.
Steps $s_t$ do not have to be uniformly equal, and can either be a hyperparameter or learned.
Similarly, other conditions in $\omega_x$ can be a hyperparameter or learned to best explain the data evolution.
In the general case, we learn $s_t$ and $\omega_x$ and parameterize all coefficients of an \mnnrep~$\alpha=[c_{i, t, x}, \phi_{k,t, x}, b_{t, x}, s_t, \omega_x]$
with a standard neural network $f_\theta(x)$ (see~\eqref{eq:mnn-1}). 
Coefficients $\alpha$ are obtained with a single forward pass for all discrete times $t=1, ..., n$.

\paragraph{Time-invariant Coefficients and Universal ODEs. \ }
In applications, we are often interested in discovering simpler universal governing ODEs with coefficients that are either \emph{time-independent} ($c_{i, x}, \phi_{k, x}$), e.g., with autonomous ODEs, or that are \emph{shared across inputs} ($c_{i, t}, \phi_{k, t}$).
For instance, we might want to automatically discover the general governing equation of planetary motions that apply to all astronomical objects.
Then, we simply drop the time dependency from the coefficients and share them dataset-wide $X=\{x^1, ..., x^B\}$ \vspace*{-0.3cm}
\begin{equation}
\mathcal{U}_x: \; \sum_{i=0}^d c_{i}(X) u^{(i)} + \sum_{k=0}^r \phi_k(X)g_k( u,u',\ldots) = b(X). \label{eq:ode-timeinvariant}
\end{equation}
\paragraph{Complexity of the forward pass. \ } For a $n$-step discretization of a one-dimensional ODE, and a single input $x$, from equation \ref{eq:ode-general-discrete} with $r$ non-linear terms requires specifying of $d+r+n+1$ parameters \emph{per time step}:
on the left-hand, $d+1$ parameters to specify the values of $c_{i, t, x}$ for orders $i=0, ..., d$ (including the 0-th order for the function evaluation), $r$ parameters $\phi_{k, t, x}$ for nonlinear terms, and $n-1$ parameters for the step sizes $s_t$; on the right-hand side, one parameter for $b_{t, x}$.
We also specify any possible initial conditions for the first time step $t=1$ up to order $d-1$, 
In practice, we use sparse matrix methods for large problems and can solve large systems efficiently. Other than estimating coefficients $\alpha$ and solving the \mnnrep~$F$, the whole forward pass is like with standard neural networks.

\paragraph{Training challenges. \ } During training, we need to compute gradients $\frac{\partial{f}}{d \alpha}, \frac{\partial{f}}{d \theta}$ through the ODE solver.
We compute these gradients per layer and perform regular backpropagation, as with standard neural networks.
The caveat is that both the forward and backward passes require a significant amount of computation for solving systems of ODEs \emph{en masse} and computing the gradients.
We thus ideally want an efficient, neural-friendly ODE solver. 

One option is general-purpose ODE solvers such as Runge-Kutta methods, which are inefficient for our case.
First, they are \emph{sequential}, thus the gradient computations are recurrent.
Second, for batches of independent ODEs general-purpose ODE solvers require independent computations.
We address both problems in the next section.

\section{Neural Relaxed LP ODE Solver}
\label{sec:method}

We present the \emph{Neural Relaxed Linear Programming (NeuRLP) solver}, a novel, efficient, and parallel algorithm for solving batches of independent ODEs.
In section \ref{sec:qp-ode-solving}, we show how we can solve linear ODEs with differentiable quadratic programming with equality constraints, motivated by a proposal~\citep{young1961linear} for representing linear ODEs as linear programs.
In section~\ref{sec:kkt}, we explain how this is, in practice, done efficiently by solving a KKT system for the forward and the backward pass.
In section~\ref{sec:nonlinear-ode}, we extend the solver for \emph{non-linear} ODEs, which is not possible solely with linear programs.
In section~\ref{sec:error-bounds}, we prove error bounds for the NeuRLP solver and show they are comparable to Euler solvers.
We analyze in section~\ref{sec:complexity} the theoretical computational and memory complexity of the solver. 

The NeuRLP solver is differentiable, GPU parallelizable for large ODE systems, supports multiple inputs in a mini-batch for hundreds of discrete times $t=1,...,n$, learnable step sizes $s_t$, and learnable initial conditions $\omega_x$, significantly improving efficiency compared to traditional sequential ODE solvers.
We compare with standard ODE solvers in section \ref{sec:ablations} including Runge-Kutta (RK4) from popular software packages, specifically \texttt{scipy} and \texttt{torchdiffeq}.

\subsection{Linear ODEs as Linear Programs}
\label{sec:qp-ode-solving}

We start with discretized linear ODEs ignoring the nonlinear terms $g_k$ in~\eqref{eq:ode-general-discrete}, that is $\sum_{i=0}^d c_{i, t, x} u^{(i)}_t = b_{t, x}, \; \text{s.t. \;}  [u_{t=1}, u_{t=1}', ...]=\omega_x$.
We reintroduce the nonlinear terms $g_k$ in section~\ref{sec:nonlinear-ode}.

As shown by \citet{young1961linear}, one can solve linear ODEs by solving corresponding (dual) linear programs of the form
\begin{align}
\begin{array}{r@{}rll}
\text{minimize }&  \displaystyle &\delta^\top z\\
\text{subject to }&  &A z \ge \beta ,
\end{array}\label{eq:dual-lp}
\end{align}
where $z$ is the variable that we optimize for and $A \in \mathbb{R}^{m\times n_v}$ and $\beta \in \mathbb{R}^m$ represent the (inequality or equality) constraints and $\delta\in \mathbb{R}^{n_v}$ represents the cost of each variable. In the following subsections, we detail the form and intuition of the different parts and variables of the linear program, that is the constraints and the optimization objective.

\subsubsection{Constraints}

Core to the linear program in ~\eqref{eq:dual-lp} are the (in)equality constraints $Az \geq \beta$.
We have three types of constraints: the equality constraints that define the ODE itself, initial value constraints, and smoothness constraints for the solution of the linear program.

\textbf{ODE equation constraints} specify that at each time step $t$ the left-hand side of the discretized ODE is equal to the right-hand side, \emph{e.g.,} for a second-order ODE,
\begin{align}
c_{2,t}u''_{t} + c_{1,t}u'_{t} + c_{0,t} u_t = b_t, \forall t\in \{1, \dots, n\}.
\label{eq:eq-constraints}
\end{align}

\textbf{Initial-value constraints} specify constraints on the function or its derivatives for the initial conditions at $t=1$, \emph{e.g.,}, that they have to be equal to 0,
\begin{align}
u_1 =0,\;\; u'_1 = 0 \label{eq:iv-constraint}
\end{align}

\textbf{Smoothness constraints} control how smooth the discretization in ~\eqref{eq:ode-general-discrete} of the continuous ODE in ~\eqref{eq:ode-general}.
In other words, the smoothness constraints make sure the solutions of the linear program to the function and derivative values at each time step are $\epsilon$-close in neighboring locations.
We determine the values in neighboring locations by Taylor approximations up to error $\epsilon$.
We define one Taylor approximation for the forward-time evolution of the ODE, $t:1  \rightarrow n$, and one for the backward-time, $t:n  \rightarrow 1$.
If we are interested in a second-order ODE for instance, we have as Taylor expansions:

\vspace{-0.6cm}
\begin{small}
\begin{align}
\begin{rcases}
| u_t + s_tu'_t + \frac{1}{2}s_t^2 u''_t - u_{t+1} | &\le \epsilon\\
 |s_tu'_t + \frac{1}{2}s_t^2 u''_t - s_tu'_{t+1} | &\le \epsilon \label{eq:taylor-1}
\;\; \end{rcases} &\;\; \text{Forward-time} \\
\begin{rcases}
 |u_t - s_tu'_t + \frac{1}{2}s_t^2 u''_t - u_{t-1} | &\le \epsilon\\
|-s_tu'_t + \frac{1}{2}s_t^2 u''_t + s_tu'_{t-1}|  &\le\epsilon\label{eq:taylor-2}
\;\; \end{rcases} &\;\; \text{Backward-time}
\end{align}
\end{small}
with $\epsilon \ge 0$.
For higher-order ODEs, we simply include to the Taylor expansions the additional smoothness constraints for the higher-order derivatives too.
Note that in equations~\ref{eq:taylor-1}-~\ref{eq:taylor-2} the coefficients of the derivatives are $s_t$ rather than $c_{\cdot}$ because they correspond to the Taylor expansions of the function in neighboring locations.

\textbf{Defining $z$.} To describe how we transfer all the above constraints to $A$ and $b$, we must first explain what goes into the variable $z$ that we will be solving for with our linear program.
In $z$, we introduce three types of variables.
First, we introduce per time step $t \in \{1,\ldots, n\}$ one variable $z_{t}^{0}$ that corresponds to the value of the function at time $t$, that is $u_t$.
Second, we introduce per time step $t \in \{1,\ldots, n\}$ one variable $z_{t}^{i}$ that corresponds to the value of the $i$-th function derivative at time $t$ for all derivative orders, that is $u_t^{(i)}, i=1, ..., d$.
Third, we introduce a single scalar variable $\epsilon$ shared for all time steps that corresponds to the error of the Taylor approximation for all function values and derivatives.
All in all, we have that $z=[z_{t}^{i}, \epsilon], t=1, ..., n, \text{and} \; i=0, ..., d$, whereby $i=0$ refers to the function value (0-order derivative).

\textbf{Defining $A, b$. \ }
In total, we have $m$ constraints (one per row) and $n_v$ variables.
We rewrite the constraints so that the terms with the variables appear on the left-hand side of the constraint and everything else on the right-hand side.
The variable coefficients then are the elements in the matrix $A \in \mathbb{R}^{m\times n_v}$.
The right-hand sides of the rewritten constraints are collected in $b$.
To clarify, the step sizes only appear in the matrix A and not in the variables z.

\textbf{Optimization objective.}
The objective of the linear program is to minimize the smoothness error $\epsilon$.
Solving the linear program, we obtain in $z$ the function values and derivatives that satisfy all the ODE equality and inequality constraints, including $\epsilon$-smoothness.
Furthermore, we also obtain a value for the minimized error $\epsilon$.

\subsection{Efficient Quadratic relaxation}

Solving ODEs using the LP method inside neural networks has three main obstacles: 1) the solutions to the LP are not continuously differentiable \citep{wilder_melding_2018} with respect to the variables $A, b, c$ that interest us and 2) solving linear programs is generally done using specialized solvers that do not take advantage of GPU parallelization and are too inefficient for neural networks applications, and 3) The matrices $A$ are highly sparse where dense methods for solving and computing gradients (such as from \citet{amos2017optnet}) are infeasible for large problems.

We can avoid the non-differentiability of linear programs by including a diagonal convex quadratic term \citep{wilder_melding_2018} as a regularization term, converting inequalities into equalities by slack variables and removing non-negativity constraints \citep{pervez2023differentiable} to obtain an equality-constrained quadratic program,
\begin{align}
\begin{array}{r@{}rll}
\text{minimize }&  \displaystyle &\frac{1}{2} z^\top G z + \delta^\top z\\
\text{subject to }&  &\tilde{A} z = \beta + \xi,
\end{array}\label{eq:eq-qp}
\end{align}
where $G=\gamma I, \gamma \in \mathbb{R}$ is a multiple of the identity for a relaxation parameter $\gamma$ and $\xi$ are slack variables.
Importantly, equality-constrained quadratic programs can be directly and very efficiently solved in parallel on GPU \citep{pervez2023differentiable}.
This is why we rewrite inequalities as equalities using slack variables.
Although with equalities only we lose the ability to  explicitly encode non-negativity constraints, we mitigate this by regularization making sure that solutions remain bounded.

\subsection{Efficient forward and backward computations}
\label{sec:kkt}

\paragraph{Forward propagation and solving the quadratic program.}
We can solve the quadratic program directly with well-known techniques \citep{wright1999numerical}, namely by simplifying and solving the following KKT system for some $\lambda \in \mathbb{R}^m$,
\begin{equation}
\begin{bmatrix}
G & A^\top\\
A & 0
\end{bmatrix}
\begin{bmatrix}
-z\\
\lambda
\end{bmatrix}
= 
\begin{bmatrix}
\delta\\
-\beta
\end{bmatrix}\label{eq:kkt_optimality}
\end{equation}
For smaller problems, we can solve this system efficiently using a dense Cholesky factorization. 
For larger problems, we use an indirect conjugate gradient method to solve the KKT system using only \emph{sparse} matrix computations.
Both methods are performed batch parallel on GPU.

\paragraph{Backward propagation and gradients computation.\ }
In the backward pass, we need to update the ODE coefficients in the constraint matrix $A$ and $\beta$.
We obtain the gradient relative to constraint matrix $A$ by computing $\nabla_A \ell(A) = \nabla_A z \nabla_z \ell(z)$, where $\ell(.)$ is our loss function and $z$ is a solution of the quadratic program.

We can compute the individual gradients using already established techniques for differentiable optimization \cite{amos2017optnet} with the addition of computing sparse gradients only for the constraint matrix $A$.
Briefly, computing the gradient requires solving the system \eqref{eq:kkt_optimality} for with a right-hand side containing the incoming gradient $g$:
\begin{equation}
-
\begin{bmatrix}
G & A^\top\\
A & 0
\end{bmatrix}
\begin{bmatrix}
d_z\\
d_\lambda
\end{bmatrix}
= 
\begin{bmatrix}
g\\
0
\end{bmatrix}.\end{equation}
The gradient $\nabla_A \ell(A)$ can then be computed by first solving for $d_z, d_\lambda$ and then computing $d_\lambda z^\top + \lambda d_z^\top$ \citep{amos2017optnet}.
In general, this would produce a dense gradient matrix, which is very memory inefficient for sparse $A$.
We avoid this by computing gradients only for the non-zero terms of $A$ by computing sparse outer products.

\subsection{Nonlinear ODEs}
\label{sec:nonlinear-ode}

The standalone solver described above works for linear ODEs.
When combined with neural networks, we can extend the approach to nonlinear ODEs by combining solving with learning.

For each non-linear ODEs term $g_k$, we add an extra variable $\nu_{k,t}$ with coefficients $\phi_{k,t,x}$ corresponding to the non-linear term to our linear program.

We rewrite our nonlinear ODE in ~\eqref{eq:ode-general-discrete-constraints} as
\vspace{-0.3cm}
\begin{align}
\sum_{i=0}^d c_{i, t, x} u^{(i)}_t + \sum_{k=0}^r \phi_{k,t,x}\nu_{k, t} &= b_{t, x} \label{eq:ode-general-discrete-expanded} \\
\nu_{k, t} = g_k(t, u_t,u_t',\ldots), &\;k=0, ..., r \label{eq:ode-general-discrete-aux-expanded}\\
\text{s.t. \;}  [u_{t=1}, u_{t=1}', ...]&=\omega_x \label{eq:ode-general-discrete-constraints-expanded},
\end{align}
that is, for each nonlinear term $k$ and for every time step $t$ we also add in $z$ an auxiliary variable $\nu_{k, t}$.
Additionally, we include derivative variables $\nu_{k, t}^{i}$ that are part of the Taylor approximations to ensure smoothness.
We then solve the linear part of the above ODE, that is equation~\ref{eq:ode-general-discrete-expanded} subject to ~\ref{eq:ode-general-discrete-constraints-expanded} with the linear programs we described in the previous subsections.
Further, we convert the nonlinear part in equation~\ref{eq:ode-general-discrete-aux-expanded} to a loss term $\big( \nu_{k, t} - g_k(t, u_t,u_t',\ldots) \big)^2$, which is added to the loss function of the neural network.
With the extra losses, we learn the parameters $\phi$ such that $\nu_k$ is close to the required non-linear function of the solution.

\paragraph{Nonlinear ODEs for Discovery.\ } When building MNN models for governing equation discovery, we incorporate nonlinear ODEs using a set of predefined basis functions $\{\theta_i\}$, such as the polynomial basis functions \citep{brunton2016discovering}, to build an equation of the form
\begin{equation}
    \frac{d}{dt} u(t;x) = \sum_i^k \theta_i(u'_x(t))
\end{equation}
The input $u'_x(t)$ to the basis functions are generated by a neural network with input $x$ as $u'_x = \text{NN}(x)$, where $u'_x$ (and possibly $x$) depends on time $t$.
To ensure that this is a proper nonlinear ODE we add a consistency term to the loss function to minimize the squared loss $(u(t;x) -u'_x(t))^2$.
This ensures that the basis input and ODE solution are close.

\subsection{Error bounds}
\label{sec:error-bounds}
We take the example of solving a second order linear ODE $c_2 u'' + c_1 u' + c_0 u = b$
over $n$ steps for a fixed step size $s$.
We show in \ref{appendix:error-analysis} that under reasonable assumptions, namely $\frac{c_0}{c_2}$ is $O(\frac{1}{s^2})$ and $\frac{c_1}{c_2}$ is $O(\frac{1}{s})$, the error over $n$ steps is bounded by $O(s^2)$. 
This bound is for a second order approximation and is comparable to the Euler method over $n$ steps.
The error can be improved by taking smaller step sizes (which we can learn) and higher-order approximation.

\subsection{Complexity}
\label{sec:complexity}

The computational and memory complexity of MNNs is determined by the size of the time grid $n$, and the order $d$ of the ODEs to be generated.
The last layer of $f$ outputs $n\times(d+2)$ ODE parameters.
This means that the memory required to store the coefficients can be large depending on the grid size and dimension.
The main computational effort in solving the system \eqref{eq:kkt_optimality} for a batch of ODEs, which we do by a Cholesky factorization for small problems or sparse conjugate gradient for large ones. 
Cholesky factorization has complexity cubic in $nd$ while conjugate gradient has quadratic complexity.

\subsection{Numerical validation of the solver}
\label{sec:ablations}

\paragraph{Benchmarking against RK4 from \texttt{scipy} and \texttt{torchdiffeq}.}
We compare with traditional ODE solvers on second- and third-order linear ODEs with constant coefficients from the \texttt{scipy} package. 
For a \emph{learning} comparison, we also compare with the RK4 solver with the adjoint method from the \texttt{torchdiffeq} package on the benchmark task of fitting long and noisy sinusoidal functions of varying lengths.
The quantitative and qualitative results in Appendix~\ref{sec:further-experiments} show that NeuRLP is comparable to standard solvers on the linear ODE-solving task.
On the \emph{fitting} task NeurLP significantly improves upon the baseline and is \emph{about 200x faster} with a lower MSE loss than the \texttt{torchdiffeq} baseline for 1000 steps.\\

\paragraph{NeuRLP can learn time steps.}
Unlike traditional solvers, NeuRLP can learn the discretization grid for learning and solving ODEs, becoming adaptively finer in regions where the fit is poor.
We validate this on fitting a damped sinusoid, see results in figure \ref{fig:solver-steps}, where we begin with a uniform grid and with steps becoming denser in regions with bad fit.

\section{Related Work}
\label{sec:related}

\paragraph{Neural Dynamical Systems.}
In terms of data-driven modeling of dynamical systems with differential equations, MNNs are related to Neural ODEs \citep{chen2018neural}. 
\looseness=-1With neural ODEs the model can be seen as the forward evolution of a differential equation.
With MNNs a set of ODEs are first generated and then solved in a single layer for a specified number of time steps.
Another important difference is that with Neural ODEs the learned equation is implicit and there is a single ODE for modeling the system.
MNNs, on the other hand, explicitly generate the dynamical equation that governs the evolution of the input datum with potential for analysis and interpretation.
Variations such as augmentation \citep{dupont2019augmented} and second-order ODEs \citep{norcliffe2020second} overcome some of the limitations. 
Universal differential equations \citep{rackauckas2020universal} can also be seen as generalization of Neural ODEs.
\looseness=-1MNNs generate a family of linear ODEs, one per initial state, with arbitrary order which makes them very flexible and enables non-linear modeling.
\citet{doi:10.1137/21M1411433} consider Neural ODE applications for control.

\paragraph{Neural PDE Solvers.}
Recently the use of Deep Learning to improve the speed and generalization of PDE solving has gained significant interest, collecting training data by solving PDEs for known initial conditions and testing for unseen initial conditions.
Fourier Neural Operator \citep{li2020fourier} uses the Fourier transform to focus significant frequencies to model PDEs with super-resolution support. 
\citet{brandstetter2021message} investigate PDE solver properties and methods for improving rollout stability.
\citet{brandstetter2022lie} consider Lie-group augmentations for improving neural operators.
MNNs show that Neural PDE solvers can be built solely with a fast and parallel ODE solver such as NeuRLP and with a performance that is close to specialized methods without special tricks.

\paragraph{Discovery.} 
\looseness=-1MNNs are also related to discovery methods for physical mechanisms with observed data.
SINDy \citep{brunton2016discovering} discovers governing equations from time series data using a pre-defined set of basis functions combined with sparsity-promoting techniques.
A number of subsequent works have extend the basis method improve robusts, PDE discovery, parameterized pattern formation etc., \citep{kaheman2020sindy, rudy2017data, nicolaou2022data}.
An advantage of MNNs is that they can handle larger amounts of data than shallow methods like SINDy. 
Other approaches to discover physical mechanisms are Physics-informed networks (PINNs) and universal differential equations \citep{rackauckas2020universal, raissi2018deep}, where we assume the general form of the equation of some phenomenon, we posit the PDE operator as a neural network  and optimize a loss that enables a solution of the unknown parameters.
In contrast MNNs parameterize a family of ODEs by deep networks and the solution is obtained by a specialized solver.

\paragraph{ODE Solvers.}
\looseness=-1Traditional methods for solving ODEs involve numerical techniques such as finite difference approximations and Runge-Kutta algorithms. 
The linear programming approach to numerical solution of linear ordinary and partial differential equations was originally proposed in \citet{young1961linear} (also see \citet{rabinowitz_applications_1968}). 
MNNs require fast and GPU parallel solution to a large number of independent but simple linear ODEs for which general purpose solvers would be too slow.
For this we revisit the linear program approach to solving ODEs, convert it to an equality constrained quadratic program for fast batch solving and resort to differentiable optimization methods \citep{amos2017optnet, barratt_differentiability_2019, wilder_melding_2018} for differentiating through the solver.

\section{Example Applications in Scientific ML}
\label{sec:experiments}

To probe the versatility of Mechanistic Neural Networks, we benchmark them in \emph{five different settings} from scientific machine learning applications for dynamical systems.
Due to the vast heterogeneity of the problems, different benchmarks and machine learning methods have prevailed as golden standards.
Per setting, we describe the state-of-the-art and a way to use Mechanistic Layers for the task. 
In the appendix, we provide a complete description of each application and ablations, visual explanations and additional results.
We share source code for our method.

\subsection{Discovery of Governing Equations}
\label{sec:mnn-discovery}

\looseness=-1\paragraph{Problem.} Often, the goal is to discover underlying laws governing the data in addition to making accurate predictions.
This is especially for applications in sciences, with great interest in physics, fluid dynamics \citep{loiseau2018constrained}, and material science \citep{alves2022data}. 

\paragraph{Gold standard} in discovering governing equations is SINDy \citep{brunton2016discovering, rudy2017data}.
SINDy models the problem as linear regression on a library of candidate nonlinear basis functions $\Theta(x)$, e.g., constant, polynomial or trigonometric ones, such that the equation discovery corresponds to the best-fitting linear combination with coefficients $\Xi$.
SINDy is fundamentally constrained to problems where the governing equations are linear combinations of simpler terms, being a \textit{linear} combination of (nonlinear) basis functions.

\paragraph{Mechanistic NNs for discovering equations.}
Similar to \cite{brunton2016discovering}, we model nonlinear ODEs $\frac{d}{dt} \mathbf x(t) = \mathbf F(\mathbf x(t))$, from ~\eqref{eq:mnn-2} with polynomial basis but followed by a further nonlinear transformation depending on the problem.
In section \ref{app:experiment-discovery} in the appendix, we provide a precise description of the model architecture and the training setting.\\
\paragraph{Experiment.}
We experiment with
the following ODE systems:
(1) the Lorenz system, and
(2) ODEs with complex nonlinear function of the form $\frac{d}{dt} \mathbf x(t) = F(p(\mathbf x(t)))$, where $p$ is a polynomial and $F$ is a nonlinear function such as tanh, and
(3) ODEs with rational function derivatives, $\frac{d}{dt} \mathbf x(t) = \frac{p(x)}{q(x)}$, where $p$ and $q$ are polynomials.
The latter two types cannot be modeled by the approach employed by SINDy.
Results are shown in Figure \ref{fig:discovery-nonlinear} and Figures \ref{fig:discovery-lorenz} and \ref{fig:discovery-rational} in the appendix.
For the Lorenz system which can be described as linear basis combinations, both SINDy and variants, as well as Mechanistic NNs recover the exact equations.
For complex nonlinear and rational function ODEs which requires nonlinear functions of basis combinations, SINDy exhibits poor generalization and overfits to the training domain.
\looseness=-1See appendices \ref{appendix:discovery-details}, \ref{app:experiment-discovery} for more details and discovered equations.

\begin{figure}[t]
\centering
  {{$\frac{d}{dt} \mathbf x(t) = \text{tanh}(p(\mathbf x(t)))$}\par\medskip}
\includegraphics[valign=c,width=0.48\linewidth]{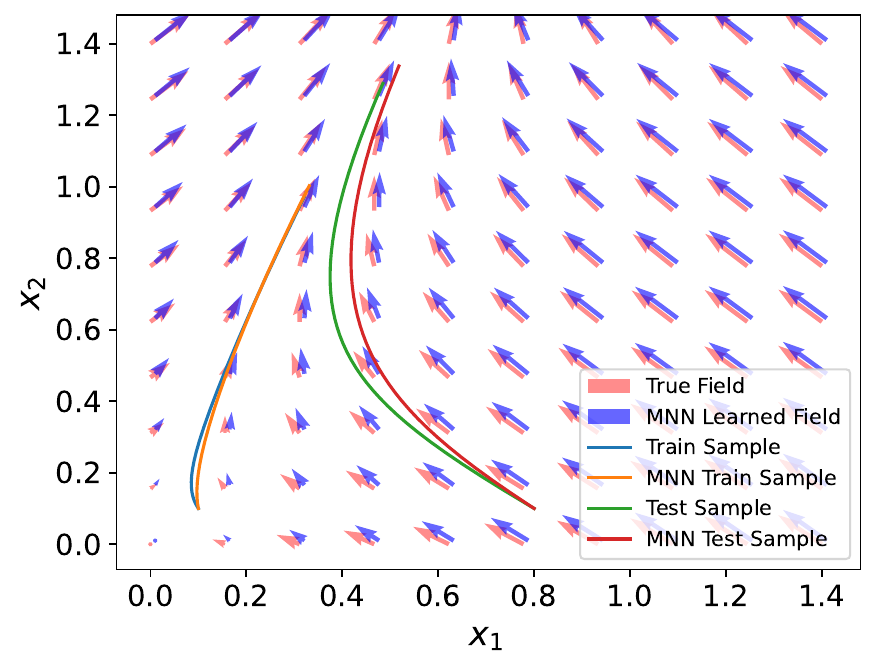}
    \includegraphics[valign=c,width=0.48\linewidth]{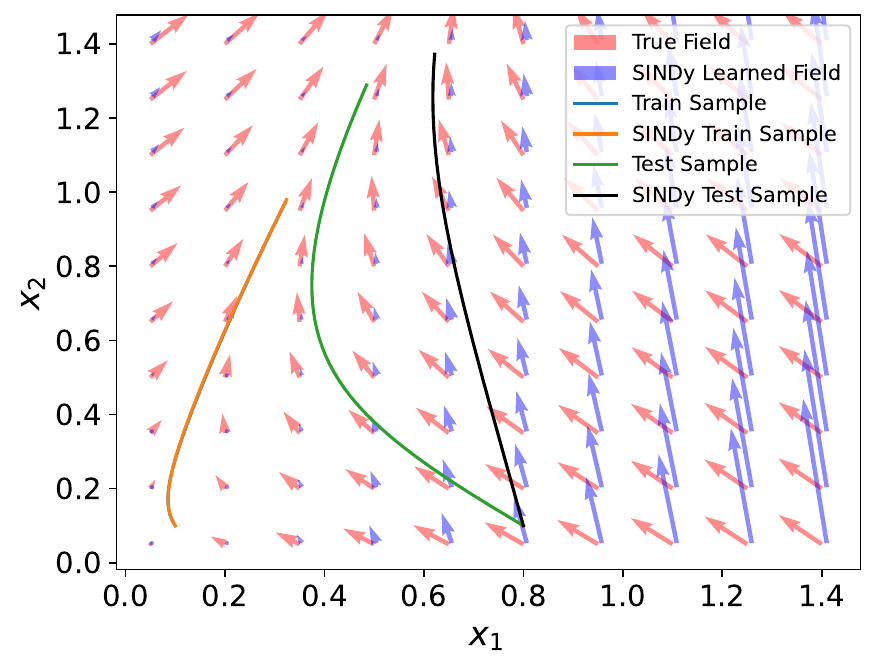}\quad

  {$\frac{d}{dt} \mathbf x(t) = \frac{p(x)}{q(x)}$\par\medskip}
    \includegraphics[width=0.48\linewidth]{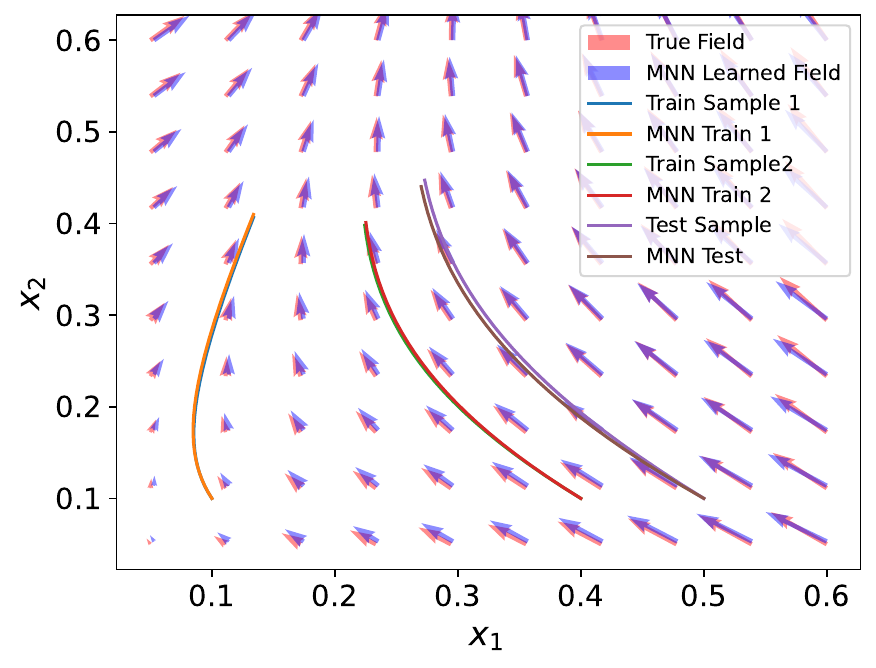}
    \includegraphics[width=0.48\linewidth]{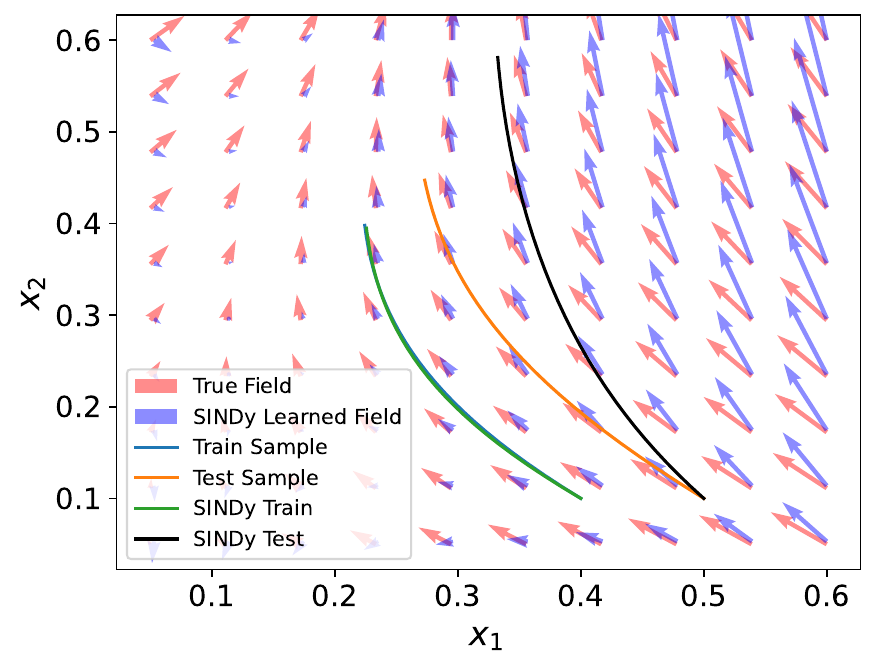}
\vskip -0.1in
  \caption{Learned ODE vector fields for Mechanistic NN (left) and SINDy (right) for ODEs with non-linear tanh and rational functions of polynomial basis.}
    \label{fig:discovery-nonlinear}
  \vskip -0.15in
\end{figure}

 \subsection{PDE Solving with Neural Networks}
\label{mnn:pde-modeling}

\paragraph{Problem.} Solving PDEs has been of tremendous interest in the past 200 years in science and engineering, from fluid dynamics to aeronautics \citep{borggaard1997pde}, weather forecasting \citep{fisher2009data}, and
Deep Learning can improve the speed and generalization of PDE solving.

\paragraph{Gold standard.} FNO \citep{li2020fourier} and Lie-group  augmented models \citep{brandstetter2022lie} are strong state-of-the-art baselines.
FNO models are deep operator architectures whose intermediate layers perform spectral operations on the input.
Lie-group augmentations for PDEs exploit that PDEs conform by definition to certain Lie symmetries to generate new training data. 

\paragraph{Mechanistic NNs for Neural PDE solving.} We adapt Mechanistic NNs from ODEs to PDEs.
For 1-d PDEs, we simply model spatial dimensions with \emph{independent} ODEs.
With a spatial dimension of 256 and prediction over 10 time steps, we learn 256 ODEs for 10 time steps each.
For 2-d PDEs we use a neural operator model with stacked MNN layers.
We provide further details of the model and training and visualizations in appendix \ref{app:pde-experiment}.

\paragraph{Experiment.} Following \citet{brandstetter2022lie}, we compare relative MSE loss using Lie-symmetry augmented ResNet, FNO and autoregressive FNO on the 1-d KdV equation (Figure~\ref{fig:kdv-pred}) and with FNO on 2-d Darcy Flow \citep{li2020fourier} (Table~\ref{app:tab:pde-darcy} in the appendix). 
We use 50 second 1-d KdV equation data and predict for 100 steps. 
We use 10 time steps of history as opposed to the baselines which use 20 time steps.
We also show visualizations for KdV prediction in Figure~\ref{fig:kdv-pred} on a 100 sec dataset.
On this heavily benchmarked setting, Mechanistic NNs are competitive with FNO and augmented models without using any specialized adaptions for stable rollout \citep{brandstetter2022lie}.

\begin{figure}
\begin{minipage}{0.55\linewidth}
\vspace{0pt}
\resizebox{0.9\linewidth}{!}{\begin{tabular}{ lcc }
\\\toprule
Method&\multicolumn{2}{c}{RMSE}\\
\midrule
&N=512 &N=256 \\
\midrule
ResNet & 0.0223 & 0.0392 \\
ResNet-LPSDA-1 & 0.0200 & 0.0284 \\
ResNet-LPSDA-2 & 0.0111 & 0.0185\\
ResNet-LPSDA-3 & 0.0155 & 0.0269\\
ResNet-LPSDA-4 & 0.0113 & 0.0184 \\
\midrule
FNO & 0.0276 & 0.0407 \\
FNO-LPSDA & 0.0055 & 0.0132 \\
FNO-AR & 0.0030 & 0.0058 \\
FNO-AR-LPSDA & 0.0010 & 0.0037 \\
\midrule
Mechanistic NN (50 sec)& 0.0039 & 0.0086 \\
\bottomrule
\end{tabular}}
\end{minipage}\begin{minipage}{0.45\linewidth}
\vspace{0pt}
\includegraphics[width=0.85\linewidth]{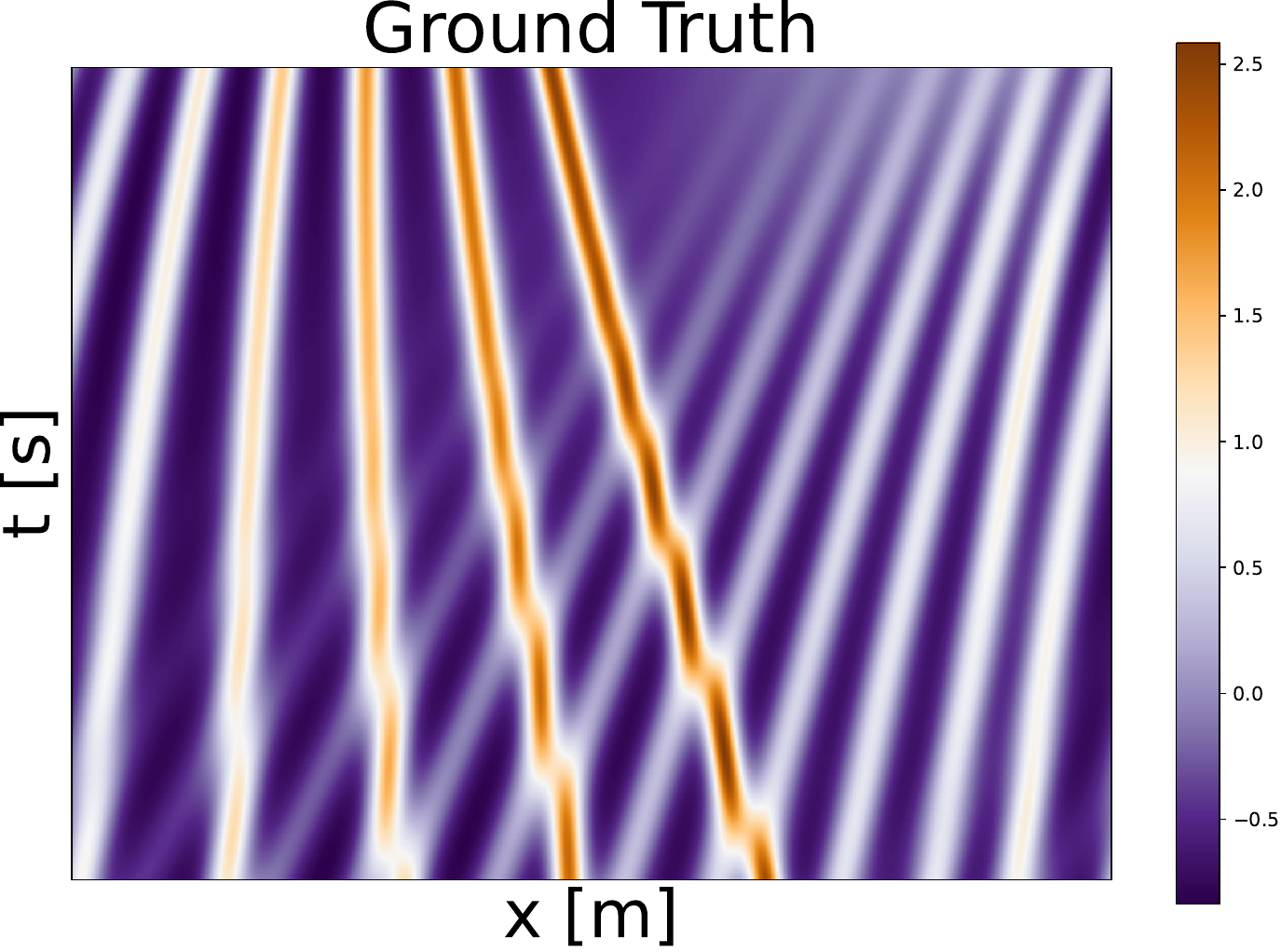}
    \includegraphics[width=0.85\linewidth]{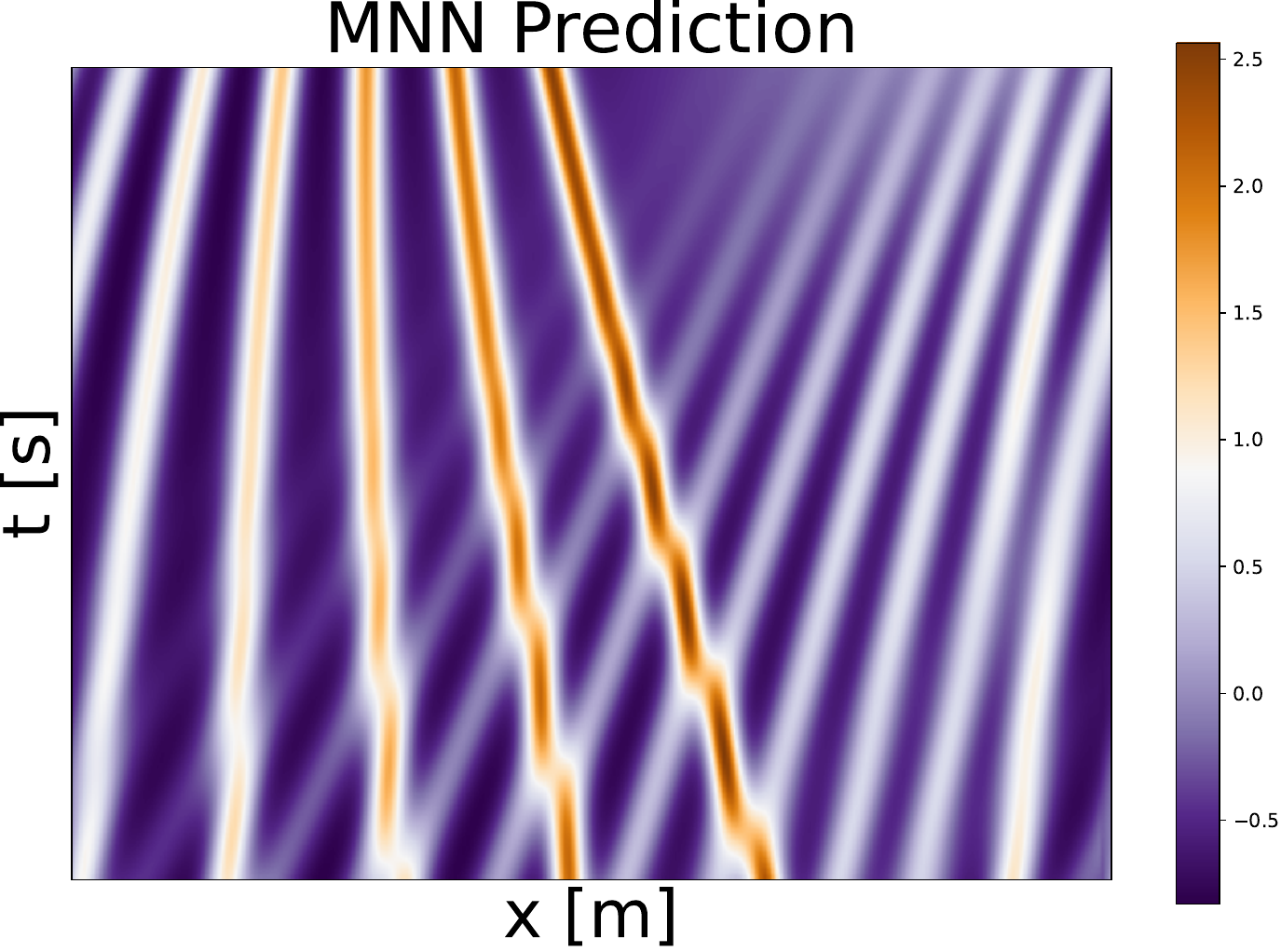}
\end{minipage}
  \caption{Solving 1d KdV\citep{brandstetter2022lie} (N train samples) (left). Comparison of ground truth and MNN prediction for the 1d KdV equation for 100 seconds (right).}
    \label{fig:kdv-pred}
\end{figure}
 \subsection{N-body Prediction}
\label{sec:mnn-nbody}
\paragraph{Problem.} N-body problems are ubiquitous, in machine learning \citep{kipf2018neural} as well as sciences including astronomy and particle physics.
The task is to predict future locations and velocities of all $N$ bodies in the system given past observations of locations and velocities.

\paragraph{Gold standard.} One can see this as a forecasting task, thus models that forecast trajectories of arbitrary lengths are relevant.
We use Neural ODEs \citep{chen2018neural, norcliffe2020second} as a gold standard baseline.

\paragraph{Mechanistic NNs for N-body modeling.} We use a basic MNN with second-order ODEs, $\ddot{\mathbf{x}} = G(\mathbf{x}, \ddot{\mathbf{x}})$.

\begin{figure}
\centering
\begin{minipage}[t]{.4\linewidth}
\vspace{0pt}
\centering
\includegraphics[width=\linewidth]{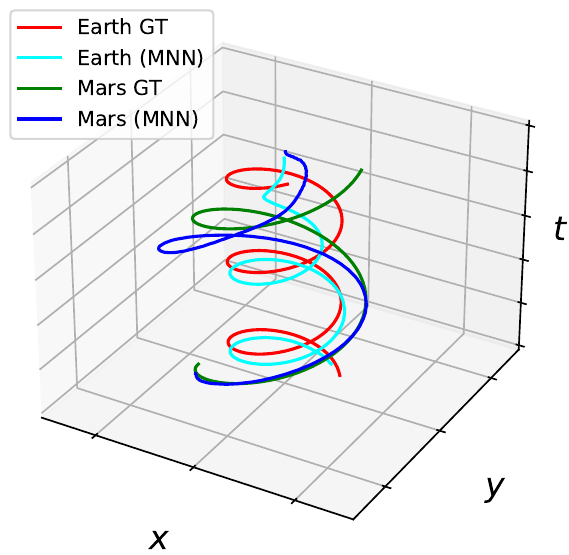}
\end{minipage}\begin{minipage}[t]{.42\linewidth}
\vspace{0pt}
\centering
\vskip 1cm
    \resizebox{0.6\linewidth}{!}{
    \begin{tabular}{ lc }
    \toprule
    Method&  Eval. MSE\\
    \midrule
ANODE & 0.0470 \\
    NODE & 0.0485 \\
    SONODE & 12.200 \\
    \midrule
    MNN & 0.0034\\
    \bottomrule
    \end{tabular}}
\end{minipage}
  \caption{Ephemerides experiment predictions for orbits of Earth, Mars (left) for 1000 days (2000 steps) and eval loss (right). Showing x,y coordinates with time for visualization.}
  \label{fig:ephemerides}
\vskip -0.2in
\end{figure}

\paragraph{Experiment.}
We use planetary ephemerides data from the JPL Horizons database for solar system dynamics \citep{2015IAUGA..2256293G}.
The data is positions and velocities for the 25 largest bodies in the solar system from 1980 to 2015 with a step size of 12 hours.
We use the first 70\% of the data for training and the rest for evaluation.
At training the MNN model predicts the next 50 steps given 50 input steps.
At testing we rollout predictions for 2000 steps given the starting 50 steps.
See prediction rollouts for Earth and Mars in and evaluation losses in Figure \ref{fig:ephemerides}. 
Mechanistic NNs improve Neural ODEs significantly by at least a factor of 10. 

 \subsection{Discovery of Physical Parameters}
\label{sec:mnn-discovery-physical}

\paragraph{Problem.} 
\looseness=-1Often, the problem is not to discover the governing equations in a system but the most fitting physical parameters explaining the observations.
Applications include inverse problems in dynamical systems \citep{wenk2020odin}.

\paragraph{Gold standard.}
We use second order Neural ODEs \citep{norcliffe2020second} to fit ODE models of corresponding to Newton's second Law, matching corresponding derivative coefficients to infer the physical parameters.

\paragraph{Mechanistic NNs for discovering physical parameters.} We use a second order ODE MNN with a time invariant 2nd order coefficient to match Newton's second Law. 
The force is learned by a neural networks as a function of position.

\begin{figure}[t!]
\begin{minipage}{0.5\linewidth}
  \centering
  \includegraphics[width=\linewidth]{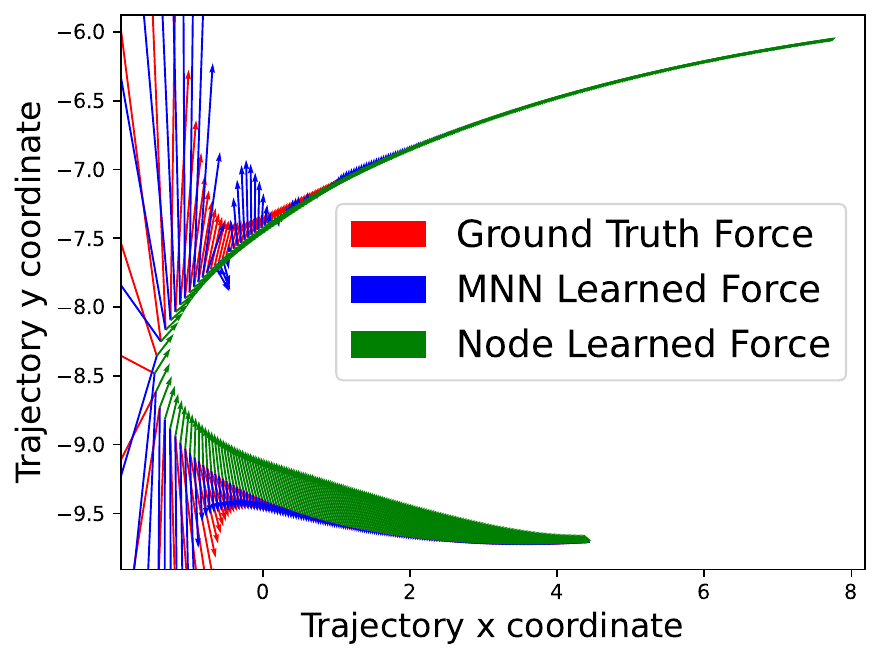}\end{minipage}\begin{minipage}{0.5\linewidth}
  \resizebox{\linewidth}{!}{
    \begin{tabular}{ lccc }
    \toprule
    Method&  Force MSE & Cosine sim. & Mass Ratio\\
    & $\downarrow$  & $\uparrow$ & GT=2\\
    \midrule
    SONODE & 879 & -0.26 & 2.11 \\
    MNN & 345 & 0.85 & 2.02\\
    \bottomrule
    \end{tabular}
  }
\end{minipage}
   \caption{Normalized true and learned force vectors during 550 steps for 2-body parameter discovery and comparison.} 
\label{fig:force-vectors-table}
\end{figure}

\paragraph{Experiment.}
We design an experiment with two bodies with masses $m_1=10, m_2=20$, distance $d$ and initial velocities $v_1, v_2$, moving under the influence of Newtonian laws, and gravitational force, $F=G \frac{m_1 m_2}{r^2}\hat{r}$,
$\hat{r}$ being the unit vector of direction of force, $G=2$ the gravitational constant.
We generate a single random train trajectory for the two bodies for 40k steps.
The physical parameters we infer are mass ratio $\frac{m_1}{m_2}$ and distribution of force values $F=[F_x, F_y]$, by combining Newton's second and third law.
We show quantitative and qualitative results in Figure~\ref{fig:force-vectors-table}.
Since forces are only determined up to a constant, to compare forces we normalize by dividing by the force at the first step.
Neural ODE and Mechanistic NNs estimate the mass ratio while MNNs perform significantly better at estimating the force distribution and Neural ODE forces often have the incorrect direction as shown by the negative cosine similarity averaged over the entire trajectory.\\

 \vspace{-5mm}
\subsection{Forecasting for time series}
\label{sec:mnn-time-series}

\paragraph{Problem.} Time-series modelling and future forecasting is a classical statistical and learning problem, usually with low-dimensional signals, like financial data or complex dynamical phenomena from sciences.

\paragraph{Gold standard.} 
We compare with Neural ODE and variants including second order and augmented Neural ODEs.

\paragraph{Mechanistic NNs for time series.} We use a basic Mechanistic NN second-order ODE for this experiment.

\paragraph{Experiment.} 
We validate on the benchmark of modeling the accelerations $a_2$ produced over time by a shaker under a wing in aircrafts \citep{norcliffe2020second}.
The model sees 1,000 past time accelerations $a_2$ and predicts the next 4000.
We show quantitative results in and qualitative results in Figure \ref{fig:plane} and \ref{appendix:airplane-vibrations} in the appendix.
The distribution of predicted $a_2$ at test time are very close to the true ones.
Mechanistic NNs are on-par with second-order ODE,   converge significantly faster, and achieve two times lower training error showing they can model complex phenomena.\\

\section{Conclusion}
\label{sec:conclusion}
This paper presented Mechanistic Neural Networks (MNNs) -- an approach for modeling complex dynamical and physical systems in terms of explicit governing mechanism. 
MNNs represent the evolution of complex dynamics in terms of families of differential equations. 
Any input or initial state can be used to compute a set of ODEs for that state using a learnable function.
This makes MNNs flexible and able to model the dynamics of complex systems.
The computational workhorse of MNNs is NeuRLP -- a new differentiable quadratic programming solver which allows a fast method for solving large batches of ODEs, allowing for efficient modeling of observable and hidden dynamics in complex systems. 
We demonstrated the effectiveness the method with experiments in diverse settings, showcasing its superiority over existing neural network approaches for modeling complex dynamical processes.

\paragraph{Limitations and future work.}
\looseness=-1In this work we have no way to measure or guarantee the identifiability of the computed ODEs, though in practice the computed equations might lie close to the true ones. Inspired by the scientific method, it would also be interesting to explore applications of MNN in active setting, with experiments performed to falsify the predictions. 
Also, in the various experiments we did not explore the model design space and better architectures and model choices can be made.
We leave all above for future work. 

\bibliographystyle{plainnat}
\bibliography{bib}
\newpage

\appendix

\section{Further Details}

\paragraph{Linear programs.}
A linear program in the primal form is specified by a linear objective and a set of  linear constraints.
\begin{mini}
{}{\displaystyle c^t x }{}{}
\addConstraint{ Ax} {=}{\;b}{}
\addConstraint{ x } {\ge}{\;0}{}
\end{mini}
where $A\in \mathbb{R}^{m\times n}$, $c\in \mathbb{R}^n$, $b\in \mathbb{R}^m$ the following specifies a linear program.
Matrix $A$ and vector $b$ define the equality constraints that the solution for $x$ must comply with.
$c^tx$ is a cost that the solution $x$ must minimize.
The linear program can also be written in dual form, 
\begin{mini}
{}{\displaystyle b^t \lambda }{}{}
\addConstraint{ A^t\lambda } {=}{\;c}{}.
\label{eq:app:dual-lp}
\end{mini}

\paragraph{Central Difference for Highest Order.} The method  proposed in \cite{young1961linear} does not add a smoothness constraint for the highest order derivative term.
In cases where a more accurate highest order term is required, we also add a central difference constraint as a smoothness condition on the highest order term.

\subsection{Error Analysis}
\label{appendix:error-analysis}
We consider the case of a second order linear ODE with an  $N$-step grid.
For simplicity we consider a fixed step size $h$, i.e., $s_t = h$.
\begin{equation}
c_2 u'' + c_1 u' + c_0 u = b,\label{error:ode}
\end{equation}
Let $u(t)$ denote the true solution with initial conditions $u(0) = r$, $u'(0) = s$.

Define 
\begin{align}
\tilde{u}_{t+1} &=  u_t + hu'_t + \frac{1}{2}h^2 u''_t, \\ 
\tilde{u}'_{t+1}  &= hu'_t + \frac{1}{2}h^2 u''_t,
\end{align}
as Taylor approximations and $\tilde{u}''_t$ is obtained by plugging the approximate values in the ODE \ref{error:ode}.

We consider the following Taylor constraints (expressions \ref{eq:taylor-1} \ref{eq:taylor-2}) for the function and its first derivative.
We use the absolute-value error inequalities for conciseness, the case for equalities is similar.
\begin{align}
| \tilde{u}_{t+1} - u_{t+1}  | &\le \epsilon \label{error:taylor-f}\\
| h\tilde{u}'_{t+1} - hu'_{t+1} | &\le \epsilon 
\label{error:taylor-fp}
 \end{align}

\paragraph{Step $t=1$.} From Taylor's theorem we have that for the first step, $t=1$,
\begin{align}
u(h) = \tilde{u}_{1} + O(h^3)\\
u'(h) = \tilde{u}'_{1} + O(h^2)
\end{align}
From \ref{error:taylor-f}, \ref{error:taylor-fp}
\begin{align}
u_1 = \tilde{u}_{1} + O(\epsilon + h^3)\\
u'_1 = \tilde{u}'_{1} + O(\epsilon/h + h^2)
\end{align}

This implies a local error at each step of $O(\epsilon + h^3)$ in $u_t$.

\paragraph{Step $t=2$.}
To estimate the error at step 2 we need to estimate the error in $u''_1$ at step 1.

For $u''_1$ we get the error by multiplying the error in $u_1$ by $\frac{c_0}{c_2}$ and that of $u'_1$ by $\frac{c_1}{c_2}$ and adding.
\begin{align}
u''_1 = \tilde{u}''_{1} + O(\frac{c_1}{c_2}(\frac{\epsilon}{h} + h^2)) + O(\frac{c_0}{c_2}(\epsilon + h^3))
\end{align}

Notice that $u''_1$ always appears with a coefficient of $h^2$.
Assuming $\frac{c0}{c2}$ is $O(\frac{1}{h^2})$ and $\frac{c1}{c2}$ is $O(\frac{1}{h})$ we have
\begin{align}
h^2 u''_1 = h^2\tilde{u}''_{1} + O(\epsilon + h^3) + O(\epsilon + h^3).
\end{align}

Each of the terms $u_1, hu'_1, h^2u''_1$ contribute an error of $O(\epsilon + h^3)$  to $u_2$ plus an additional error of $O(h^3)$ arising from the Taylor approximation and an error of $\epsilon$ arising from the inequalities \ref{error:taylor-f}, \ref{error:taylor-fp}.

\paragraph{$N$ Steps.} Proceeding similarly, after $N$ steps we get a cumulative error of $O(N(\epsilon + h^3)).$

For $\epsilon \approx h^3$ and $N \approx 1/h$, we get an error of $O(h^2)$ for $N$-steps under the assumption that $\frac{c0}{c2}$ is $O(\frac{1}{h^2})$ and $\frac{c1}{c2}$ is $O(\frac{1}{h})$.

The analysis implies that the cumulative error can become large for equations where $\frac{c0}{c2}$, $\frac{c1}{c2}$ are large.
Or, in little omega notation, $\frac{c0}{c2}$ is $\omega(\frac{1}{h^2})$ and $\frac{c1}{c2}$ is $\omega(\frac{1}{h})$ .

\subsection{Non-linear ODE Details}
In this section we give some further details regarding the formualtion of non-linear ODEs.
We illustrate with the following non-linear ODE as an example
\begin{equation}
c_2(t) u'' + c_1(t) u' + d_0(t) u^3 + d_1(t) u'^2 = b,
\end{equation}
where $u^3$ and $u'^2$ are non-linear functions of $u,u'$.

As described in Section \ref{sec:qp-ode-solving}, we create one set of variables $u_t$ for each time step $t$ for the solution $u$.
In addition we create a set of variables $\nu_{0,t}$ for $u^3$ and another set of variables $\nu_{1,t}$ for $u'^2$.
In addition we create variables $\nu'_{i,t}, \nu''_{i,t}$ for derivatives for each $i$, as in Section \ref{sec:qp-ode-solving}.

Next we build constraints. We add equation constraints for each time step as follows.
\begin{equation}
c_{2,t}u''_{t} + c_{1,t}u'_{t} + d_{0,t} \nu_{0,t} + d_{1,t} \nu_{1,t} = b_t, \forall t\in \{1, \dots, n\}.
\end{equation}
We add smoothness constraints for each $\nu_{i,t}$ in the same way as described for $u_t$ in expressions \ref{eq:taylor-1}, \ref{eq:taylor-2}.

Next we solve the quadratic program to obtain $u_t, u'_t, \nu_{0,t}, \nu_{1,t}$ in the solution.
Now we need to relate the $nu_{0,t}, nu_{1,t}$ variables to non-linear functions of the $u_t,u'_t$ variables.
For this we add the term to the loss function
\[
\frac{1}{N}\sum_t (u_t^3 -  \nu_{0,t})^2 + ({u'}_t^2 -  \nu_{1,t})^2.
\]
Figure \ref{fig:solver-non-linear} shows solving and fitting of a non-linear ODE.

\subsection{Discovery}
\label{appendix:discovery-details}
We give further details regarding the ODE discovery setup. 

The input to the basis functions is produced by a neural network from the input data.
Learnable parameters $\xi$ specify the weight of each basis function by computing $\Theta \xi$.
An arbitrary nonlinear function $g$ may be applied to $\Theta \xi$ to produce the right hand side of the differential equation as $u' = F(u) = g(\Theta\xi)$.
Given the right hand side as $F(u)$ we use the QP solver to solve the ODE to produce the solution $u$.
Given data $x$ we compute a two part loss function: The first part minimizes the MSE between $x$ and $u$ and the second part minimizes the MSE between the neural network output $f(x)$ and $x$.
Using the neural network in this way allows for a higher capacity model and allows handling noisy inputs.
The model is then trained using gradient descent.
Following \cite{brunton2016discovering} we also threshold the learned parameters $\xi$ to produce a sparse ODE solution.

\looseness=-1Vector fields for the learned systems are shown in Figure \ref{fig:discovery-nonlinear} for MNN and SINDy.
We see that although SINDy fits the training example, the directions diverge further away.
With MNN we see that the learned vector field is consistent with the ground truth far from the training example even though we use only a single trajectory.

In the following we show two examples of such cases where $F(u)$ is a rational function (a ratio of polynomials) and when $F(u)$ is a nonlinear function of $\Theta\xi$.
Moreover, unlike SINDy, MNN can learn a single governing equation from multiple trajectories each with a different initial state making MNN more flexible. 
In many situations a single trajectory sample is not enough represent  to the entire state space while multiple trajectories allow discovery of a more representative solution. 

\begin{figure}[t]
\centering
    \includegraphics[width=0.3\linewidth]{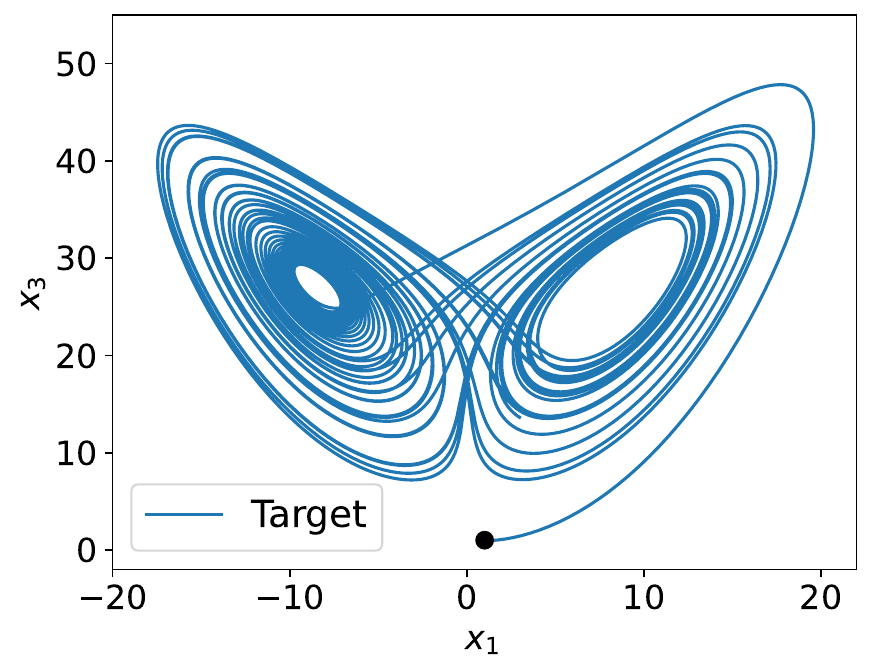}
    \includegraphics[width=0.3\linewidth]{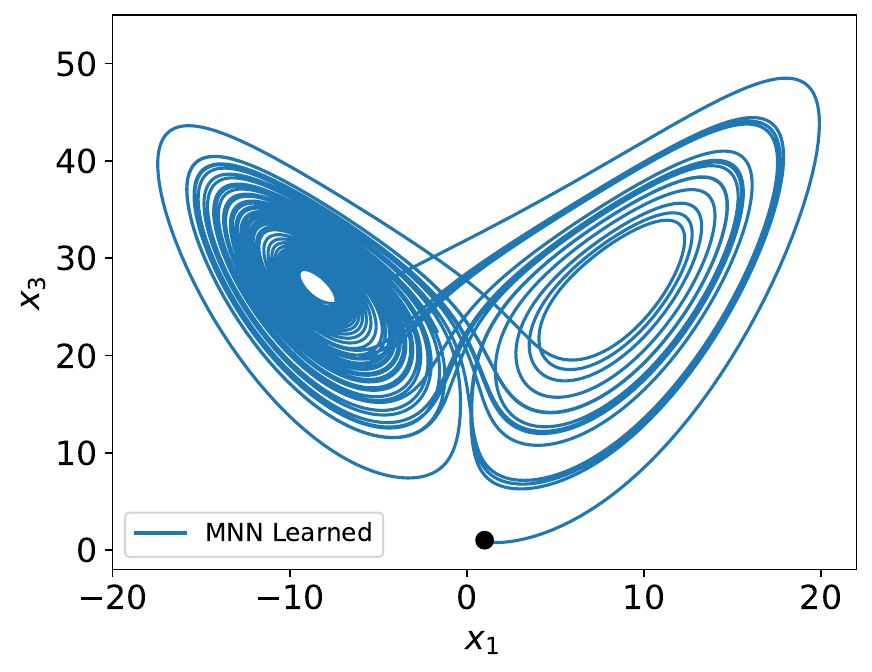}
    \includegraphics[width=0.3\linewidth]{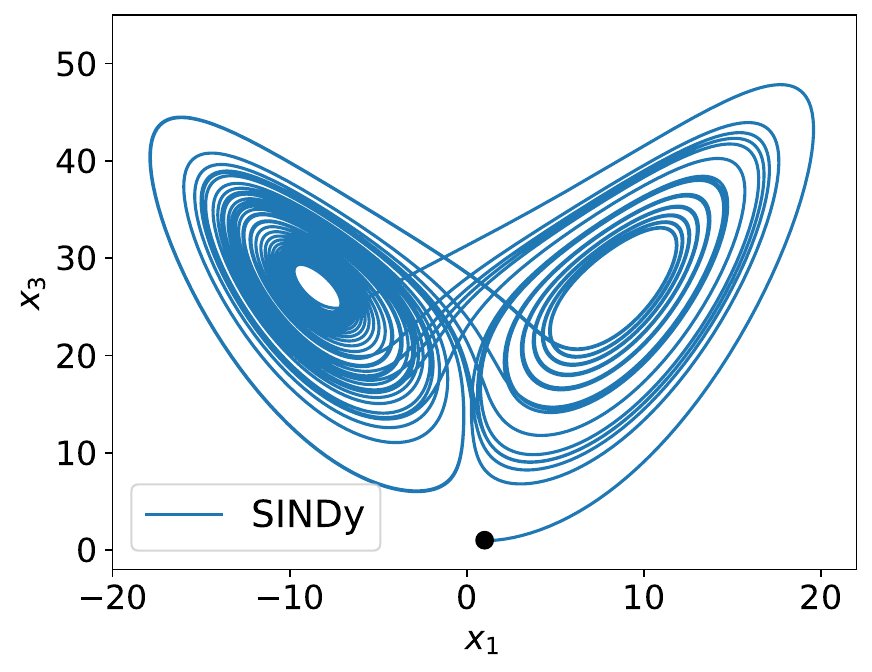}
\caption{Learned ODEs for the chaotic Lorenz system. Showing the true trajectory, the MNN learned ODE trajectory and SINDy learned ODE trajectory.} 
    \label{fig:discovery-lorenz}
  \vskip -0.15in
\end{figure}

\textbf{Planar and Lorenz System.} 
We first examine the ability of MNN equation discovery for systems where the true ODE can be exactly represented as a linear combination of polynomial basis functions.
We use a two variable planar system and the chaotic Lorenz system as examples.
Both MNN and SINDy are able to recover the planar system.
Simulation of the learned Lorenz ODE are shown in Figure \ref{fig:discovery-lorenz} for MNN and SINDy.

Next we consider ODE systems where the derivative \emph{cannot} be written as a linear combination of polynomial (or other) basis function.

\textbf{Nonlinear Function of Basis.}
First, we consider systems where the derivative is given by a nonlinear function of a polynomial.
For simplicity we assume that the nonlinear function is known.
As an example we solve the system from Figure \ref{fig:discovery-nonlinear} with the $\text{tanh}$ nonlinear function.

\looseness=-1Vector fields for the learned systems are shown in Figure \ref{fig:discovery-nonlinear} for MNN and SINDy.
We see that although SINDy fits the training example, the directions diverge further away.
With MNN we see that the learned vector field is consistent with the ground truth far from the training example even though we use only a single trajectory.

\textbf{Rational Function Derivatives.}
Second, we consider the case where the derivative is given by a rational function, i.e., $F(u)=p(u)/q(u)$, where $p$ and $q$ are polynomials.
Such functions cannot be represented by the linear combination of polynomials considered by SINDy, however such functions can be represented by MNNs by taking $p$ and $q$ to be two separate combinations of basis polynomials and dividing.
An example is shown in Figure \ref{fig:discovery-rational} in the appendix for the system
where  we see again MNNs learning much better equations compared to SINDy with a second-order polynomial basis tha overfits.
Further, by including more trajectories in the training, results improve further, see Figure \ref{fig:discovery-rational}.

We provide further details of the discovery method from Section \ref{sec:mnn-discovery}.
This method follows the SINDy \cite{brunton2016discovering} approach for discovering sparse differential equations using a library of basis functions.
Unlike SINDy, which resorts to linear regression, the MNN method uses deep neural networks and builds a non-linear model which allows modeling of a greater class of ODEs. 

The method requires a set of basis functions such as the polynomial basis functions up to some maximum degree.
Over two variables $x,y$ this is the set of functions $\{0,x,y, x^2, xy, y^2, \\xy^2, \ldots, y^d\}$ for some maximum degree $d$.
Let $k$ denote the total number of basis functions.

Next we are given some observations $X = [(x_0, y_0), (x_1, y_1), \ldots, (x_{n-1},y_{n-1})]$ for $n$ steps.
We first transform the sequence by applying an MLP to the flattened observations producing another sequence of the same shape.
\[
\tilde{X} = [(\tilde{x}_0, \tilde{y}_0), \ldots, (\tilde{x}_{n-1},\tilde{y}_{n-1})] = \text{MLP}(X) 
\]
We apply the basis functions to $\tilde{X}$ to build the basis matrix $\Theta\in \mathbb{R}^{n\times k}$.

\begin{small}
\begin{equation}
\Theta(\tilde{X}) = 
\begin{bmatrix}
1 & \tilde{x}_0 & \tilde{y}_0 & \tilde{x}_0^2 & \tilde{x}_0\tilde{y}_0 & \tilde{y}_0^2 &\ldots\\
1 & \tilde{x}_1 & \tilde{y}_1 & \tilde{x}_1^2 & \tilde{x}_1\tilde{y}_1 & \tilde{y}_1^2 &\ldots\\
\vdots & \vdots & \vdots & \vdots & \vdots & \vdots\\
1 & \tilde{x}_{n-1} & \tilde{y}_{n-1} & \tilde{x}_{n-1}^2 & \tilde{x}_{n-1}\tilde{y}_{n-1} & \tilde{y}_{n-1}^2 &\ldots
\end{bmatrix}
\end{equation}
\end{small}

Let $\xi \in \mathbb{R}^{n\times 2}$ be a set of parameters, with each column specifying the active basis functions for the corresponding variable in $[\dot{x}, \dot{y}]$.

The ODE to be discovered is then modeled as 
\begin{equation}
[\dot{x}, \dot{y}] = f(\Theta(\tilde{X})\xi) \label{eq:app:disovery-ode}
\end{equation}
where $f$ is some arbitrary differentiable function.
Note that for SINDy $\tilde{X} = X$ and $f$ is the identity function and the problem is reduced to a form of linear regression adapted to promote sparsity in $\xi$.
SINDy estimates the derivatives using finite differences with some smoothing methods.

With MNN the ODE \ref{eq:app:disovery-ode} is solved using the quadratic programming ODE solver to obtain the solution $\bar{x}_t, \bar{y}_t$ for $t \in \{0,\ldots, n-1\}$.
The loss is then computed as the MSE loss between $\tilde{x}_t, \tilde{y}_t$, $\bar{x}_t, \bar{y}_t$ and the data $x_t, y_t$.
\[
\text{loss} =\frac{1}{N}\sum_t (\tilde{x}_t - x_t)^2 + (\tilde{y}_t - y_t)^2 + (\bar{x}_t - x_t)^2 + (\bar{y}_t - y_t)^2
\]

\subsection{Physical Parameter Discovery}
We give more details about the parameter discovery setting.

We know from Newton's second and third law that $F=m \ddot{x}$, where $\ddot{x}$ is the acceleration, and $F_1=-F_2$ respectively.
By combining the two, the ratio of masses is equal $\frac{\ddot{p}_2}{\ddot{p}_1} = -\frac{m_1}{m_2}$.
To estimate, therefore, the ratio of masses $\frac{m_1}{m_2}$ we train the model to predict a differential equation $\ddot{x} = H(\mathbf{x}(t))$ for the 2-dimensional $\mathbf{x}=[x_1, x_2]$ in the \emph{xy}-plane.
For the Mechanistic NN, we constrain the library to include precisely ODE terms for the Newton laws.
The differential equation we discover with Mechanistic NNs comprises the basis functions $u_{i, j}, \ddot{u}_{i, j}$ and four coefficients $c_{i, j}$ for the $i=1, 2$ objects and the $j=1, 2$ directions in the \emph{xy}-plane.
The force $F$ is computed by neural networks satisfying the third law and superposition.

\section{Experimental Details}

\subsection{Discovery of Governing Equations}
\label{app:experiment-discovery}

\begin{figure}[h!]
\centering
    \includegraphics[height=2.5cm]{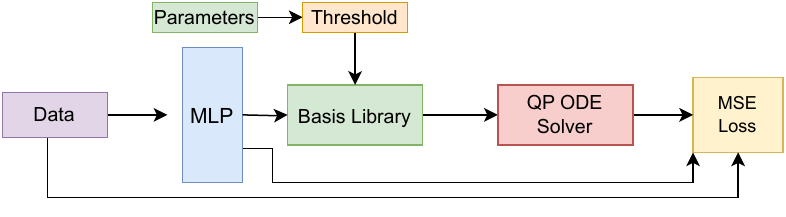}
  \vskip -0.1in
  \caption{Governing equation discovery architecture}
    \label{fig:discovery-arch}
\end{figure}

\subsubsection{Discovering governing equations of systems with rational function derivatives}

\begin{figure}[t]
\includegraphics[width=0.45\linewidth]{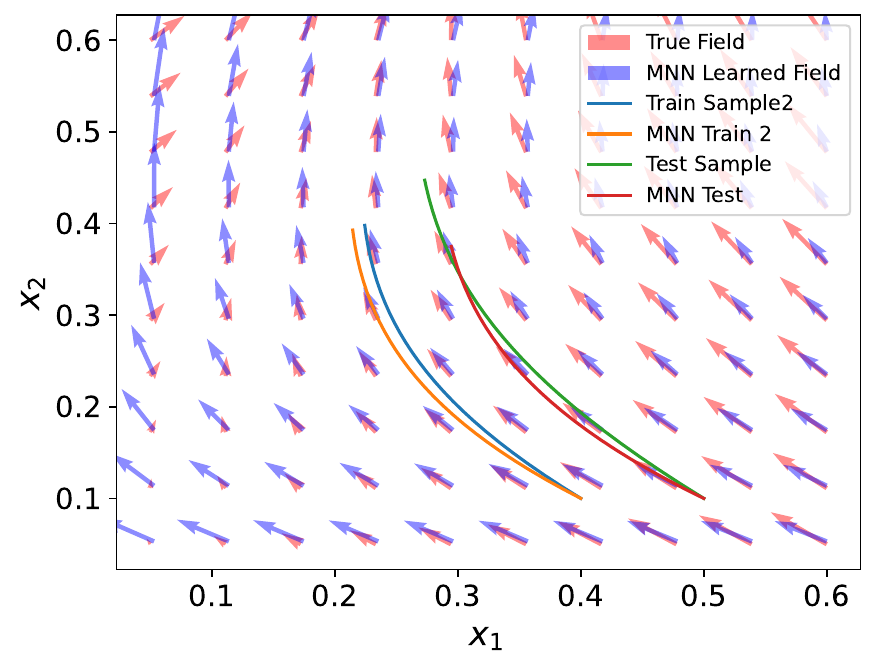}
    \includegraphics[width=0.45\linewidth]{img/discovery/rational2/mnn_rational_2_traj_test.pdf}

    \includegraphics[valign=c,width=0.45\linewidth]{img/discovery/rational2/sindy_rational_traj_test.pdf}
    \hskip 0.5cm
 \parbox[b][1cm][c]{.1\textwidth} {
 \begin{small}
 \[
\begin{aligned}
\frac{dx}{dt} &= \frac{-2x + y}{1+x^2}\\
\frac{dy}{dt} &= \frac{x + y}{1+y^2},
\end{aligned}
\]
 \end{small}
}
\caption{Learned ODE vector fields for MNN and SINDy with rational function derivatives and one and two training trajectories. MNN can handle multiple input examples. The ground truth ODE is also shown. }
    \label{fig:discovery-rational}
  \vskip -0.15in
\end{figure}

In Figure~\ref{fig:discovery-rational} we plot the vector fields learned with SINDy and with MNNs.
MNNs are considerably more accurate.

\begin{figure}[t]
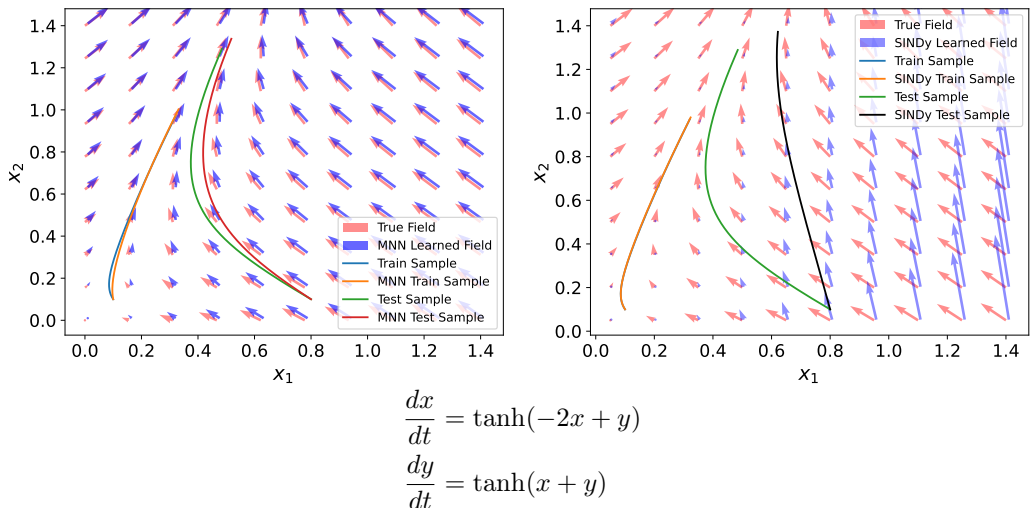

\centering
\includegraphics[valign=c,width=0.45\linewidth]{img/discovery/tanh2/mnn_tanh_traj_test.pdf}
    \includegraphics[valign=c,width=0.45\linewidth]{img/discovery/tanh2/sindy_tanh_traj_test.pdf}\quad

   \vskip 0.2cm
 \parbox[b][1cm][c]{.2\textwidth} {
 \begin{small}
 \[
\begin{aligned}
\frac{dx}{dt} &= \text{tanh}(-2x + y)\\
\frac{dy}{dt} &= \text{tanh}(x + y)\\
\end{aligned}
\]
 \end{small}
}
\caption{Learned ODE vector fields for Mechanistic NN and SINDy with non-linear tanh function  of basis combination and training  and test trajectories.} 
    \label{fig:app:discovery-nonlinear}
\end{figure}

\subsubsection{Discovered Equations.}

\textbf{MNN Lorenz} 
\begin{align*}
x' &= -10.0003x + 10.0003y\\ 
y' &= 27.9760x + -0.9934y -0.9996 xz\\
z' &= -2.6660z + 0.9995xy 
\end{align*}

\textbf{SINDy Lorenz }
\begin{align*}
x' &= -10.000 x + 10.000 y\\
y' &= 27.998 x + -1.000 y + -1.000 x z\\
z' &= -2.667 z + 1.000 x y
\end{align*}

\textbf{MNN Non-linear}
\begin{align*}
x' &= \text{tanh}( -0.7314x + 0.5545y +\\&\quad -1.2524x^2 + -0.1511xy + 0.2134y^2)\\
y' &= \text{tanh}(0.9879x + 1.0005y + 0.1742x^2)
\end{align*}

\textbf{SINDy Non-linear}
\begin{align*}
x' &= -1.968 x + 0.985 y+ -0.054 x^2\\
y' &= 1.466 y + 11.892 x^2 + -5.994 x y + 0.085 y^2
\end{align*}

\textbf{MNN Rational}
\begin{align*}
x' &=\frac{ -0.9287x + 0.4386y + -1.1681x^2 + 0.3545y^2}{0.4871 + 0.8123x + 0.0984x^2 + 0.3700xy + 0.3081x^2}\\
y' &=\frac{ 0.6360x + 0.5971y + 0.3267x^2}{0.6090 + 0.7507x^2 + 0.5694y^2}
\end{align*}

\textbf{SINDy Rational} 
\begin{align*}
x' &= -1.705 x + 0.899 y + -0.318 x^2\\
y' &= -0.795 + 3.072 y + 4.777 x^2 + 6.892 x y + -4.681 y^2
\end{align*}

\subsection{Nested Spheres}
\label{sec:nested-circles}

\begin{figure}[t]
{\includegraphics[valign=c, width=\linewidth]{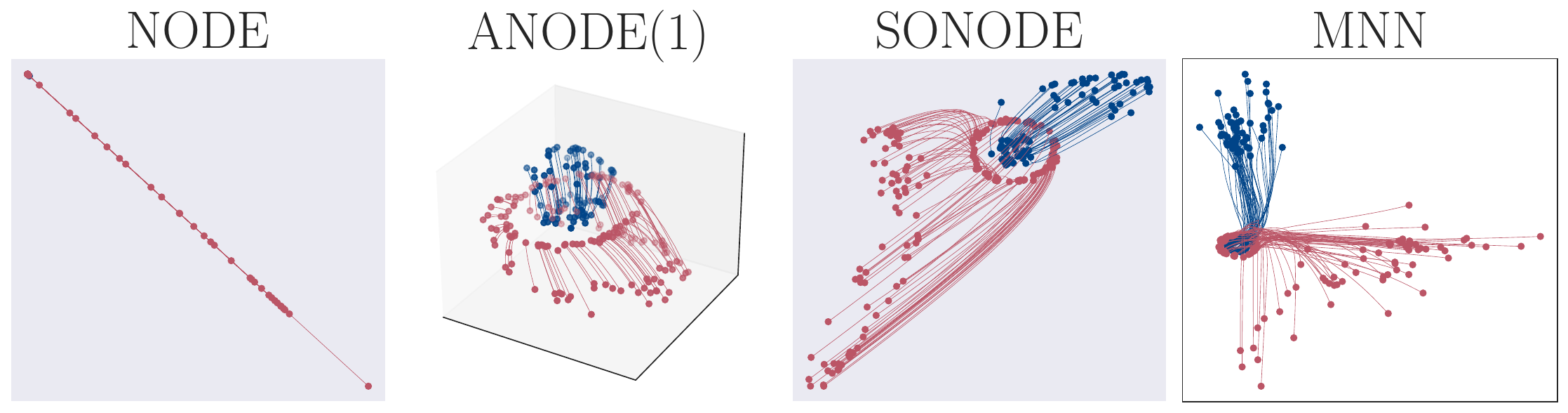}}
\caption{ Visualizing the state evolution of the learned equations $\mathcal{U}_x$ data points in nested spheres. The points from the two classes are perfectly separated despite the nested topology without requiring augmentations. }
\label{fig:nested-spheres}
\end{figure}

\looseness=-1We test MNN on the nested spheres dataset \citep{dupont2019augmented}, where we must classify each particle as one of two classes.
This task is not possible for unaugmented Neural ODEs since they are limited to differomorphisms \citep{dupont2019augmented}.
We show the results in Figure \ref{fig:nested-spheres}, including comparisons with Neural ODE \citep{chen2018neural}, Augmented Neural ODE and second-order Neural ODE \citep{norcliffe2020second}.
MNNs can comfortably classify the dataset without augmentation and can also derive a governing equation.

We use a second order ODE with coefficients computed with a single layer and the right hand side is set to 0.
We use a step size of 0.1 and length 30.
However, as we note, 5 time steps are enough for accurate classification.
The loss function is the cross entropy loss.

\subsection{Airplane Vibrations}
\label{appendix:airplane-vibrations}
\begin{figure}[t]
\includegraphics[width=\linewidth]{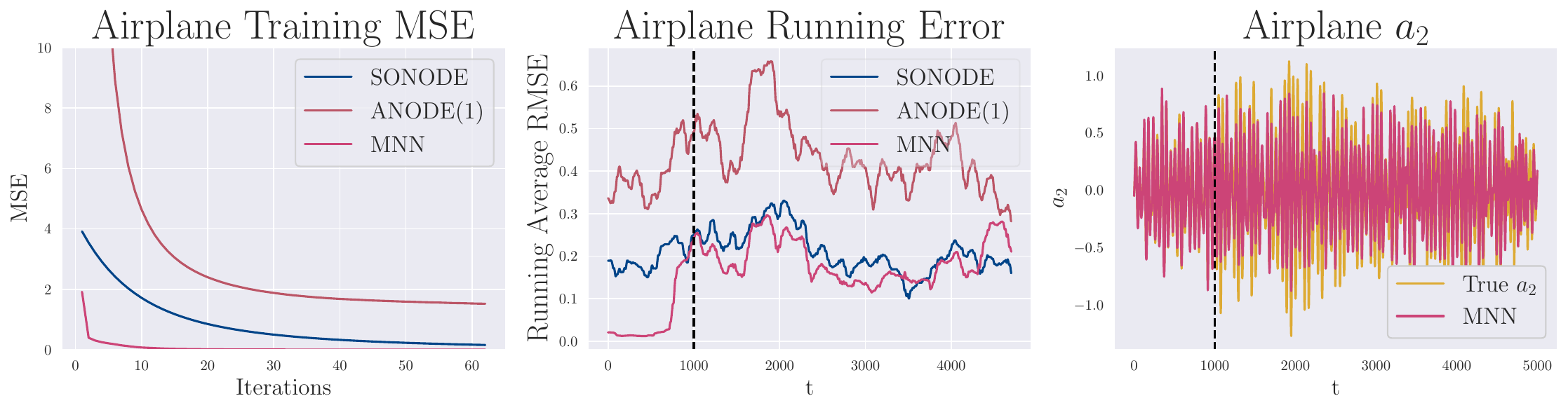}

\caption{Modeling airplane vibrations.}
    \label{fig:plane}
\end{figure}

MNNs can learn complex dynamical phenomena significantly faster than Neural ODE and second order Neural ODE.
We reproduce an experiment with a real-world aircraft benchmark dataset \citep{noel2017f, norcliffe2020second}.
In this dataset the effect of a shaker producing acceleration under a wing gives rise to acceleration $a_2$ on another point.
The task is to model acceleration $a_2$ as a function of time using the first 1000 step as training only and to predicting the next 4000 steps.
Results of the experiment are shown in Figure \ref{fig:plane}.
We compare against Augmented Neural ODE and second order Neural ODE.
MNNs are on-par with second-order ODE,   converge significantly faster in the number of training steps, and achieve two times lower training error, showcasing the capacity for modeling complex phenomena and improving with modest architectural modifications.
The predicted $a_2$ accelerations are very close to the true ones in the center-right plot.

For this experiment (Section \ref{sec:mnn-time-series}) we use an MNN with a second order ODE, step size of 0.1 and 200 steps during training.
The coefficients and constant terms are computed with MLPs with 1024 hidden units.

\subsection{Discovering Mass and Force Parameters.}
For this part of the experiment we use an MNN with a restricted ODE to match Newton's second law.
In the MNN model for this experiment, we use the same coefficient for the second derivative term for all time steps with the remaining coefficients fixed to 0, that is $c_2(t)=c$ and $c_1(t)=0, c_0(t)=0$.  
$b(t) = F_t$ corresponds to the force term which is computed by a neural network from the initial position and velocity with two hidden layers of 1024 units and Newton's second law $F_{21} = -F_{12}$. 
We use a step size of $0.01$ and run for 50 epochs.

The baseline is an SONODE designed to correspond to Newton's second and third law with an MLP for force as above.

\subsection{PDE Solving}
\label{app:pde-experiment}
\paragraph{1d Model.}
For 1d problems we use a simplest possible model of modeling the spatial dimension by independent ODEs.
We use a history of 10 time steps and predict for 9 time steps in one iteration, using the last time step as initial condition for the ODE.
During evaluation we predict and evaluate for 100 steps.
We use 3rd and 4th order ODEs.
The coefficients for ODEs, step sizes the right hand side ($b$) are computed by 1d ResNets with 10 blocks.
We use the L1 loss which we find improves rollout performance.

\begin{figure}[t]
\centering
    \includegraphics[width=0.49\linewidth]{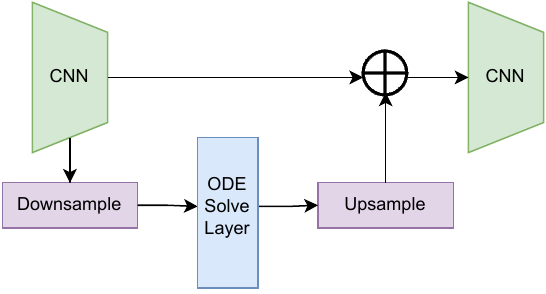}
  \vskip -0.1in
  \caption{PDE module architecture used for 2d data}
    \label{fig:pde-module}
  \vskip -0.15in
\end{figure}

\paragraph{2d Model.} In Figure~\ref{fig:pde-module} we show the MNN architecture we used to solve PDEs.
We use the 2d Darcy Flow dataset used by \cite{li2020fourier} scaled to 85x85.
The ODE is solved for 30 steps and the entire soluton trajectory is then upsampled and combined with the input features map.
The network is built by stacking three such modules together plus an input MLP layer and an output layer.

\begin{table}
\centering
    \begin{tabular}{ lc }
    \\\toprule
    Method& RMSE\\
    \midrule
    NN\cite{li2020fourier} &0.1716 \\
    FCN\cite{li2020fourier} &0.0253\\
    PCANN \cite{bhattacharya2021model} &0.0299\\
    RBM\cite{li2020fourier}& 0.0244 \\
    \midrule
    GNO \cite{li2020neural}&0.0346 \\
    LNO \cite{li2020fourier}& 0.0520 \\
    MGNO\cite{li2020multipole}& 0.0416 \\
FNO \cite{li2020fourier}& 0.0070\\
    \midrule
Mechanistic NN & {0.0065}\\
    \bottomrule
    \end{tabular}
    \caption{PDE results on 2d Darcy flow}
    \label{app:tab:pde-darcy}
\end{table}

\section{Further Experiments}
\label{sec:further-experiments}

\subsection{Validating the NeuRLP ODE Solver}
\begin{figure}[h!]
\centering
 \resizebox{8cm}{!}{\centering
    \includegraphics[width=\linewidth]{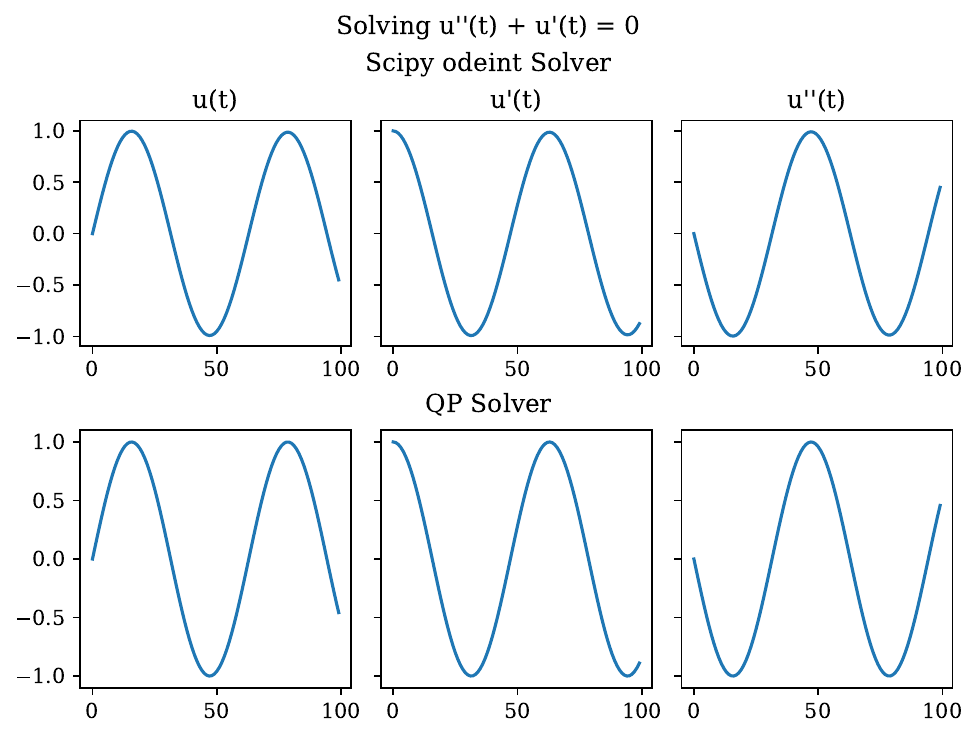}
    }

 \resizebox{8cm}{!}{\includegraphics[width=\linewidth]{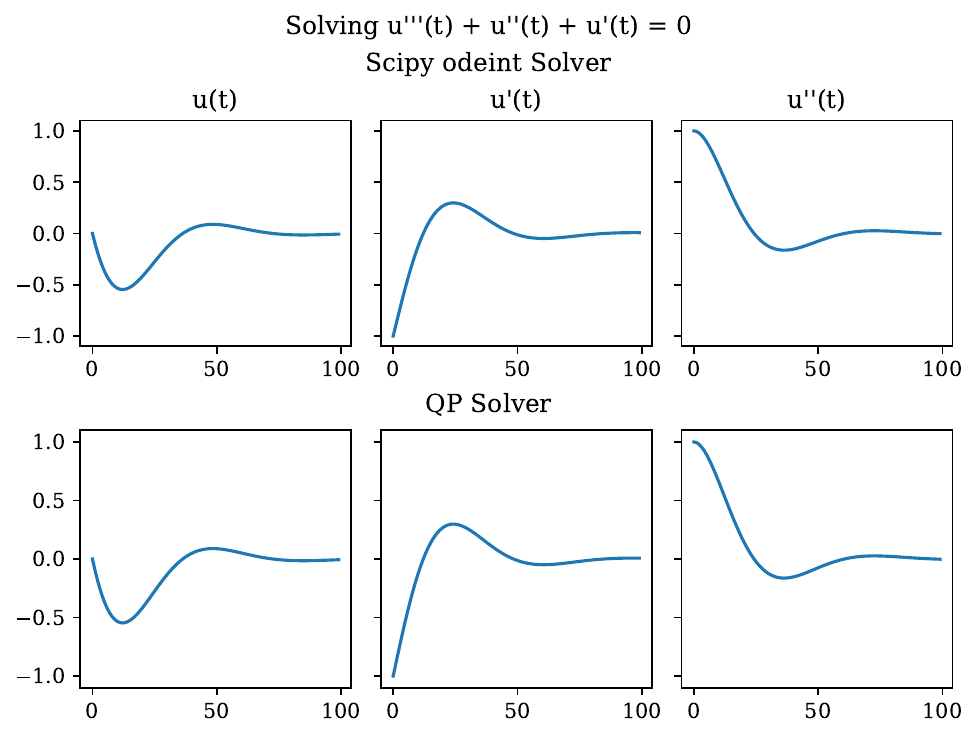}
    }
    \caption{Comparing ODE solvers on 2nd and 3rd order ODEs.}
    \label{fig:solver-comparison}
\end{figure}

First we examine whether our quadratic programming solver is able to solve linear ODEs accurately.
For simplicity we choose the following second and third order linear ODEs with constant coefficients.
\begin{align}
u'' + u &= 0\\
u''' + u'' + u' &= 0
\end{align}
For the NeuRLP solver we discretize the time axis into 100 steps with a step size of 0.1.
We compare against the ODE solver \texttt{odeint} included with the  SciPy library.
The results are shown in \ref{fig:solver-comparison} where we show the solutions, $u(t)$, for the two ODEs along with the first and second derivatives, $u'(t), u''(t)$.
The results from the two solvers are almost identical validating the quadratic programming solver.

\begin{figure}[h!]
\centering
\includegraphics[width=0.3\linewidth]{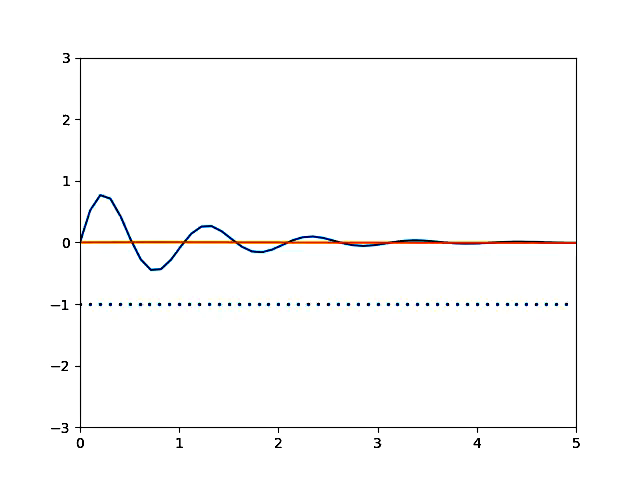}
\includegraphics[width=0.3\linewidth]{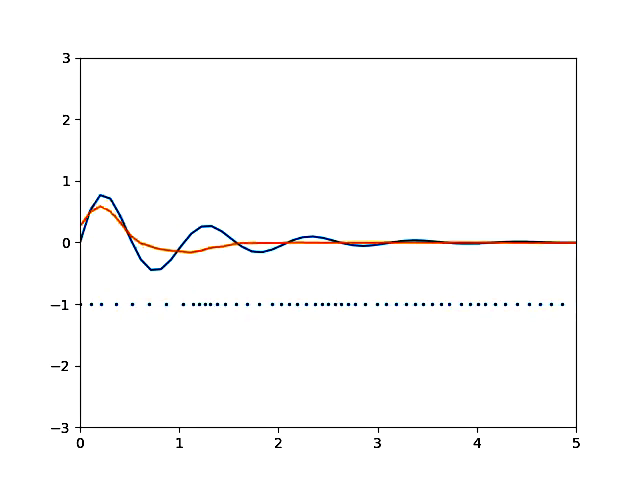}
\includegraphics[width=0.3\linewidth]{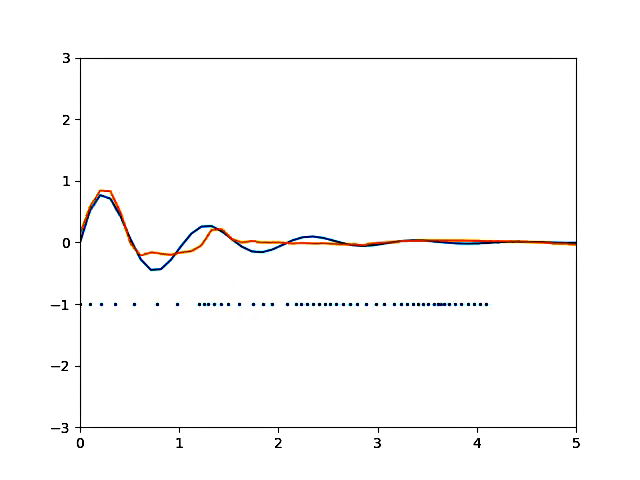}
\includegraphics[width=0.3\linewidth]{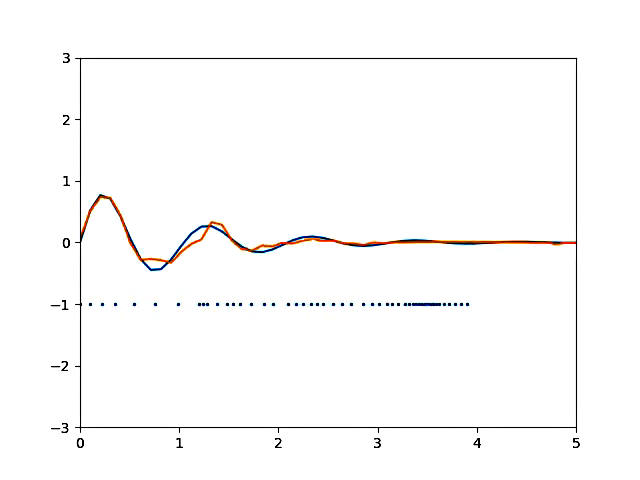}
\includegraphics[width=0.3\linewidth]{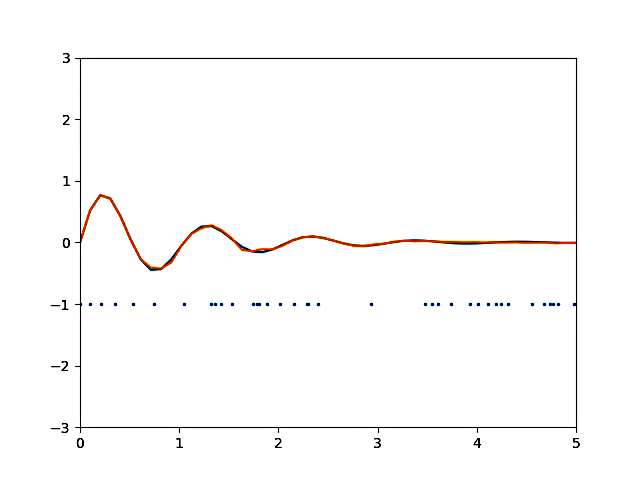}
    \caption{Demonstrating a learned grid for fitting a damped sinuoidal wave (blue curve) over the course of training. The dots show the learned grid positions. The grid generally becomes finer for regions where the fit is poorer.}
    \label{fig:solver-steps}
\end{figure}

Next we examine the ability of the solver to learn the discretization. 
We learn an ODE to model a damped sine wave where each step size is a learanable parameter initial to 0.1 and modeled as a sigmoid function.
We show the results in Figure \ref{fig:solver-steps} for a sample of training steps.
We see the step sizes varying with training and the steps generally clustered together in regions with poorer fit.

\hskip -2cm

\begin{figure}[h!]
    \centering
\hskip -1cm
\includegraphics[width=0.5\linewidth]{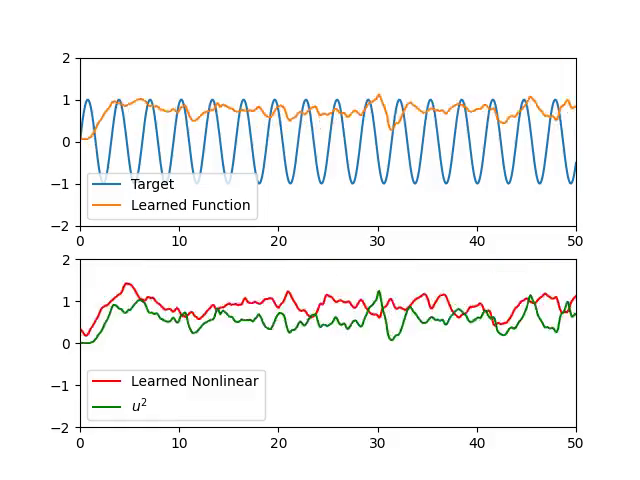}
    \hskip -0.5cm
\includegraphics[width=0.5\linewidth]{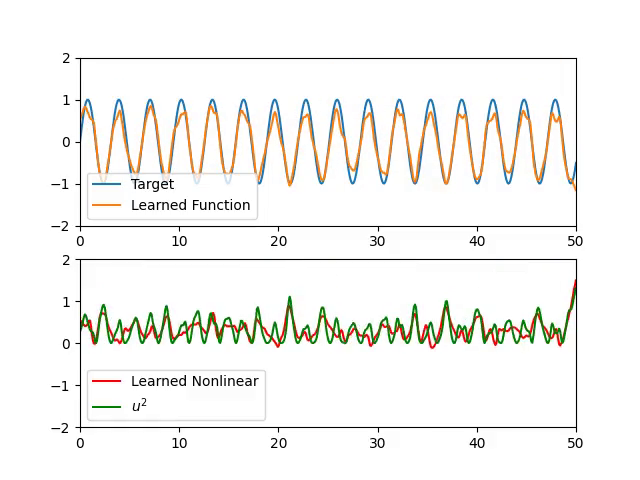}

    \hskip -2cm
\includegraphics[width=0.5\linewidth]{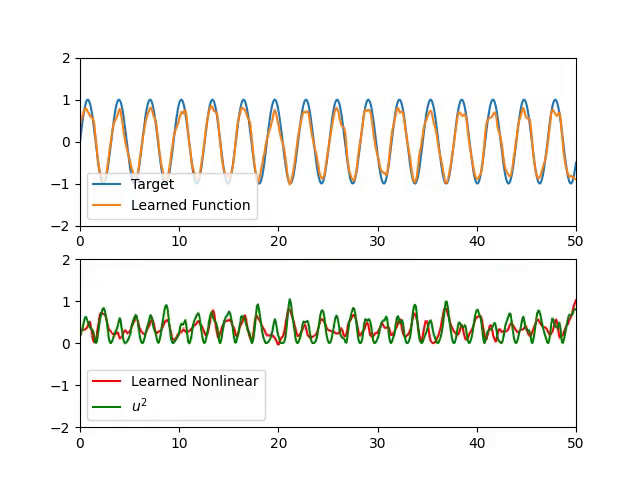}
    \hskip -0.5cm
    \includegraphics[width=0.5\linewidth]{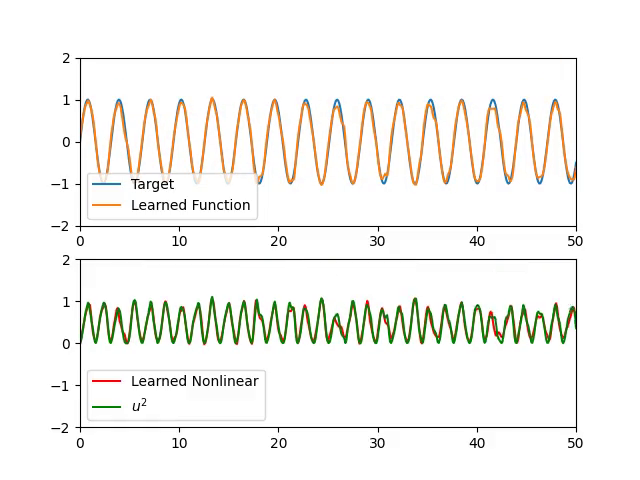}
    \hspace*{-1.5cm}
    \caption{Demonstrating fitting a sine wave with a non-linear ODE $c_2(t)y'' + c_1(t)y' + c_0(t)y + \phi(t)y^2 = 1$. The non-linear function is $y^2$ and the bottom shows the solver variable fitting the non-linear function.}
    \label{fig:solver-non-linear}
\end{figure}

Next we demonstrate a non-linear equation.
For this we introduce a variable in the QP solver for a non-linear term add a squared loss term as described in the paper.
We use the equation
$c_2(t)y'' + c_1(t)y' + c_0(t)y + \phi(t)y^2 = 1$, with time varying coefficients and fit a sine wave.
The result is in Figure \ref{fig:solver-non-linear}. 
The ODE fits the sine wave and at the same time the non-linear solver term fits the true non-linear function of the solution.

\subsection{Learning with Noisy Data}
\begin{figure}[h!]
    \centering
    \includegraphics[width=0.32\linewidth]{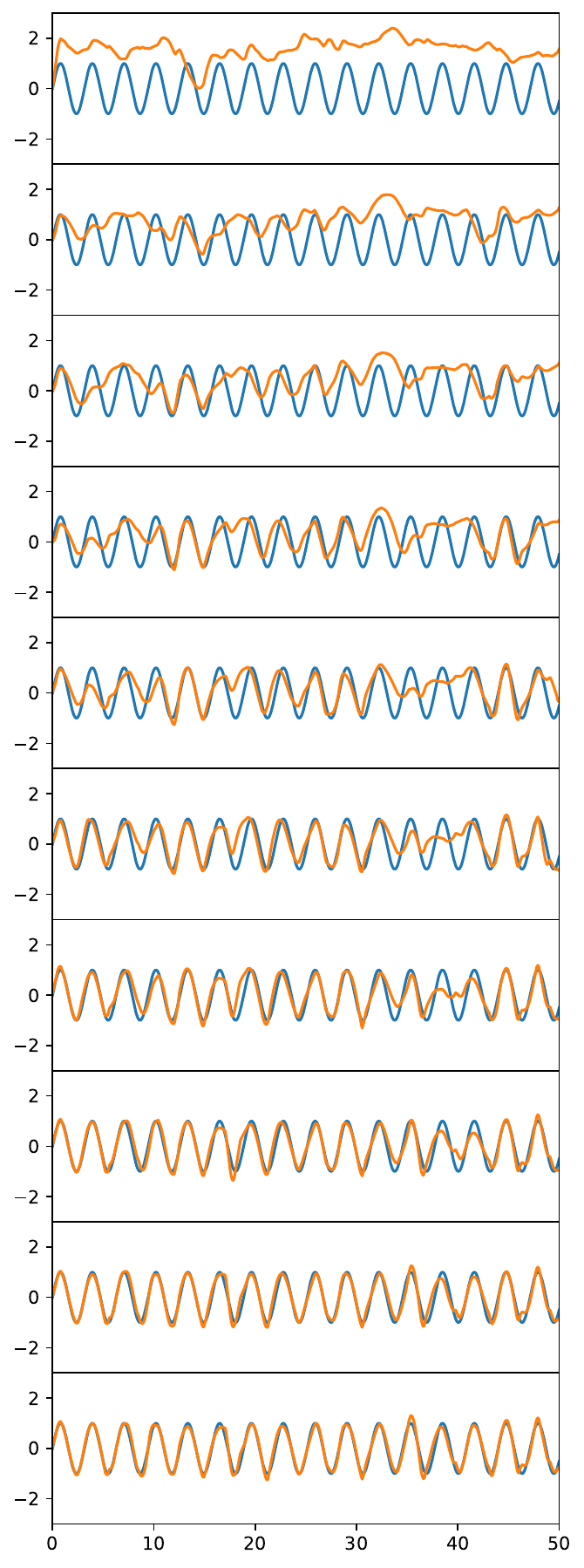}
    \includegraphics[width=0.32\linewidth]{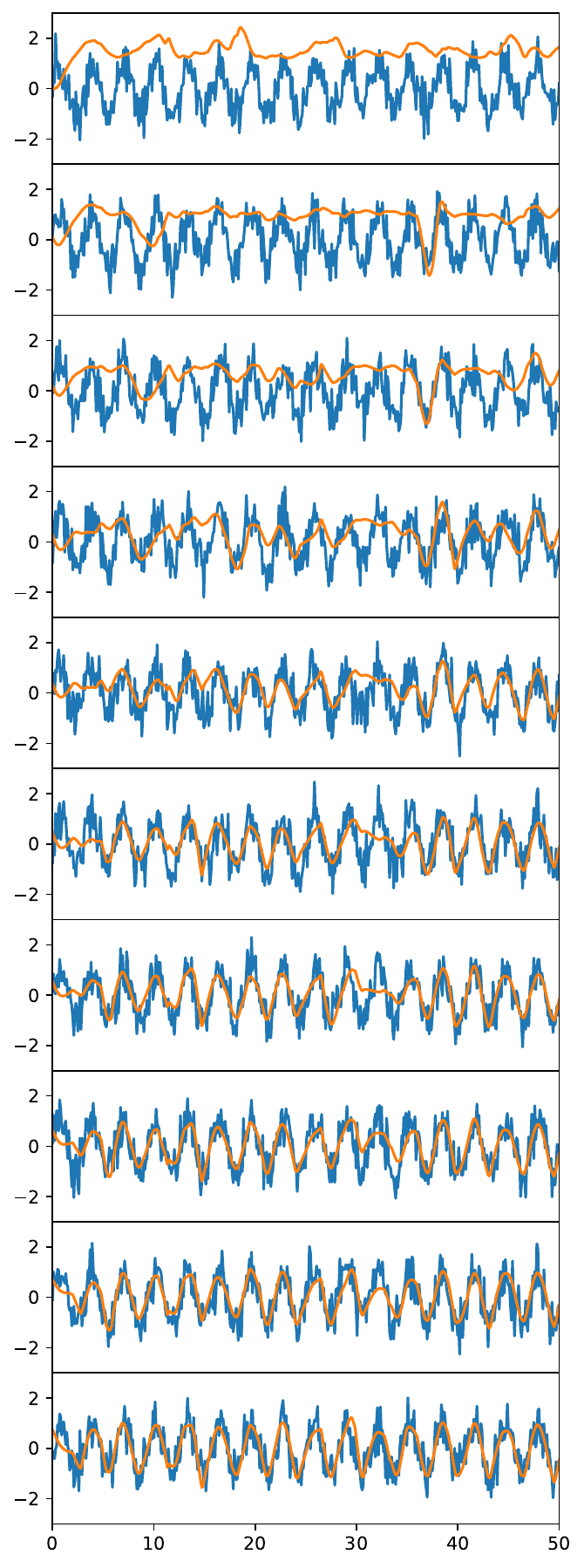}
    \includegraphics[width=0.32\linewidth]{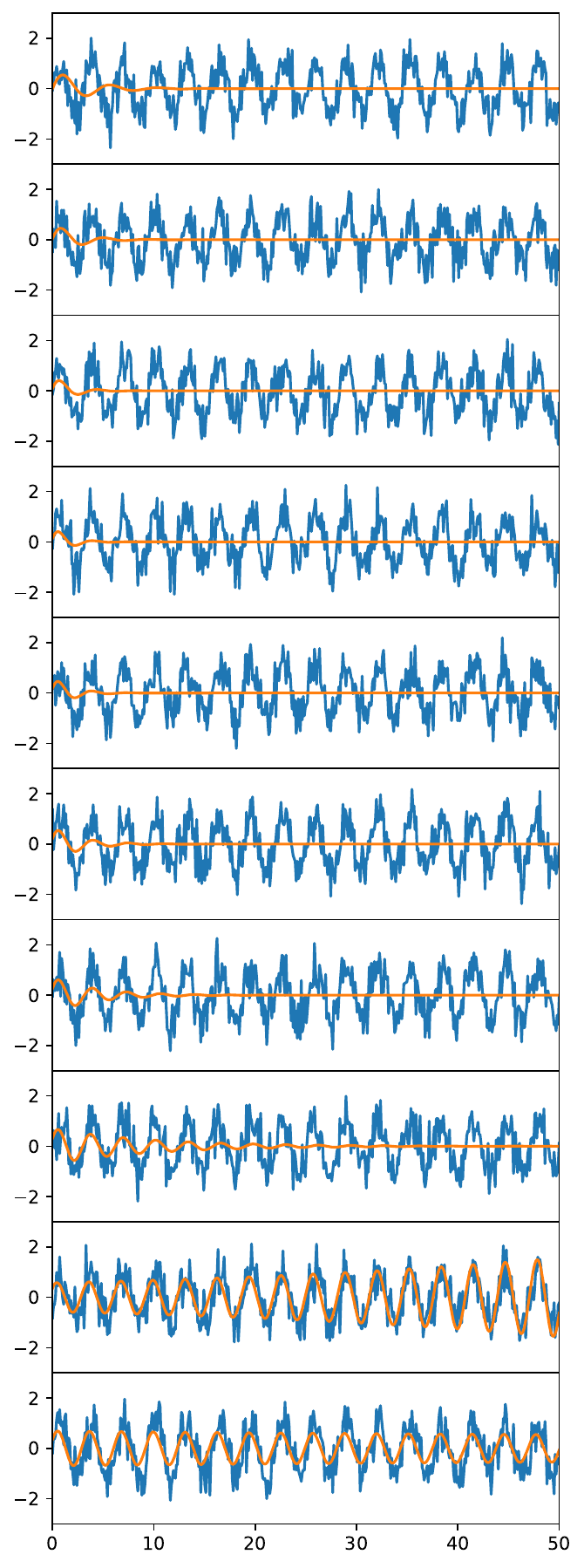}
    \caption{Learning sine waves without and with dynamically added Gaussian noise with 2nd order ODE with arbitrary coefficients (middle) and constant coefficients (right). The figure on the right corresponds to the ODE $0.92023u'' -0.00016u' +   0.228u = 0$.}
    \label{fig:robustness}
\end{figure}

We perform an simple experiment illustrate how the ODE learning method can fit ODEs to noisy data.
We generate a sine wave with dynamic Gaussian noise added during each training step.
We train two models: the first a homogeneous second order ODE with arbitrary  coefficients and the second a homogeneous second order ODE with constant coefficients.
We also train a model without noise.
The results are shown in Figure \ref{fig:robustness}.
The figures show that the method can learn an ODE in the presence of noise giving a smooth solution.
The model with constant coefficients learns the following ODE.
\[
 0.92023u'' -0.00016u' +   0.228u = 0,
 \]
 with (learned) initial conditions $u(0) = -0.031799$ and $u'(0)= 2.3657 $.

\subsection{2-Body Problem}
\begin{figure}[H]
\centering
\includegraphics[width=0.4\linewidth]{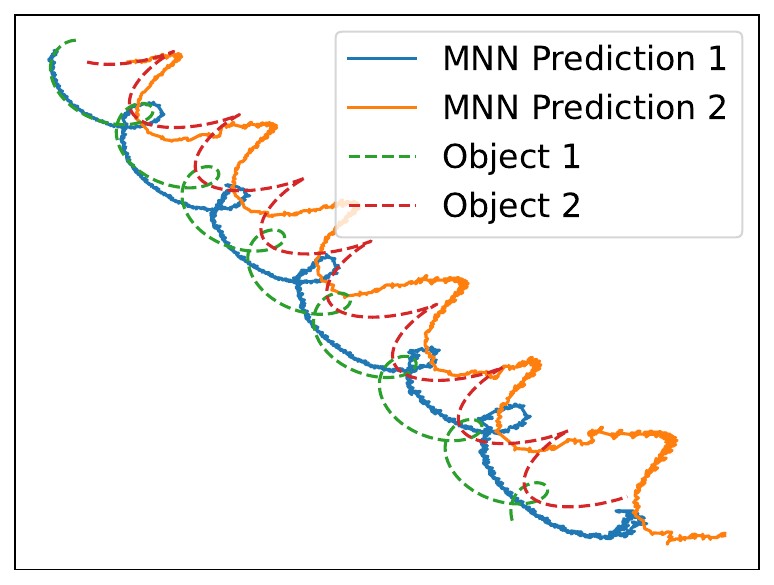}
\caption{2-body problem: Predicted orbits for MNN} 
    \label{fig:2-body-mnn-orbit}
\end{figure}

We show learned trajectories for a 2-body prediction problem with an MNN on synthetic data in Figure
\ref{fig:2-body-mnn-orbit}.
The objects are generated using the gravitation force law for 4000 steps and the first half are used for training and we predict the second half.

\subsection{Comparing RK4 with the NeuRLP solver}
\begin{figure}\begin{tabular}{cc}
\vspace{-0.2in}\\
\multicolumn{2}{c}{RK4 Solver}\\
\vspace{-0.2in}\\
  \includegraphics[width=0.4\linewidth]{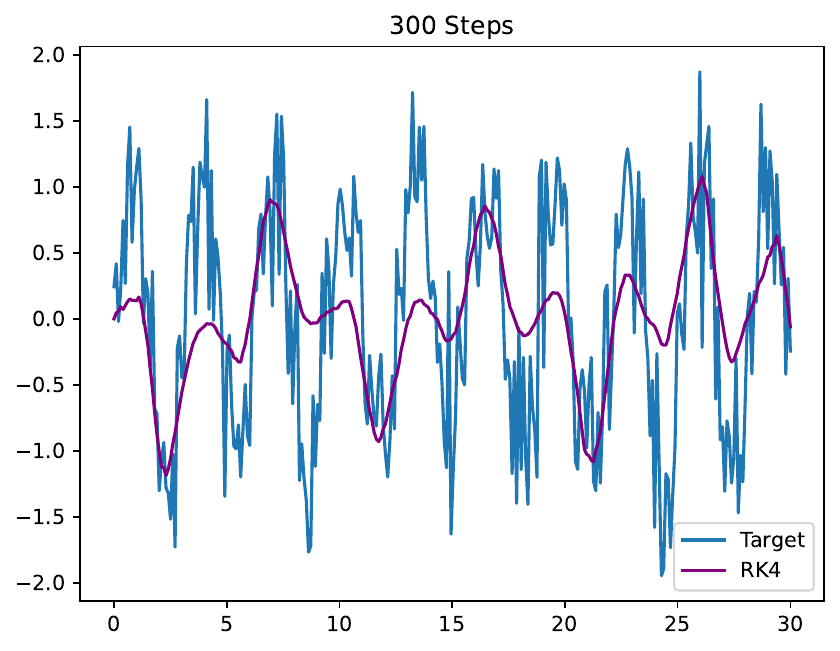}   &\includegraphics[width=0.4\linewidth]{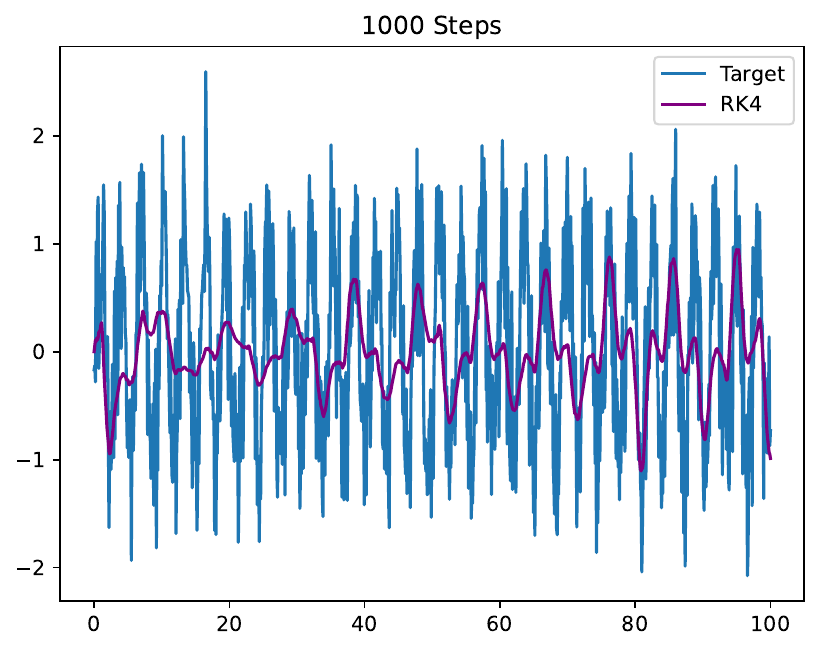}\\
\vspace{-0.2in}\\
\multicolumn{2}{c}{NeuRLP Solver}\\
\vspace{-0.2in}\\
  \includegraphics[width=0.4\linewidth]{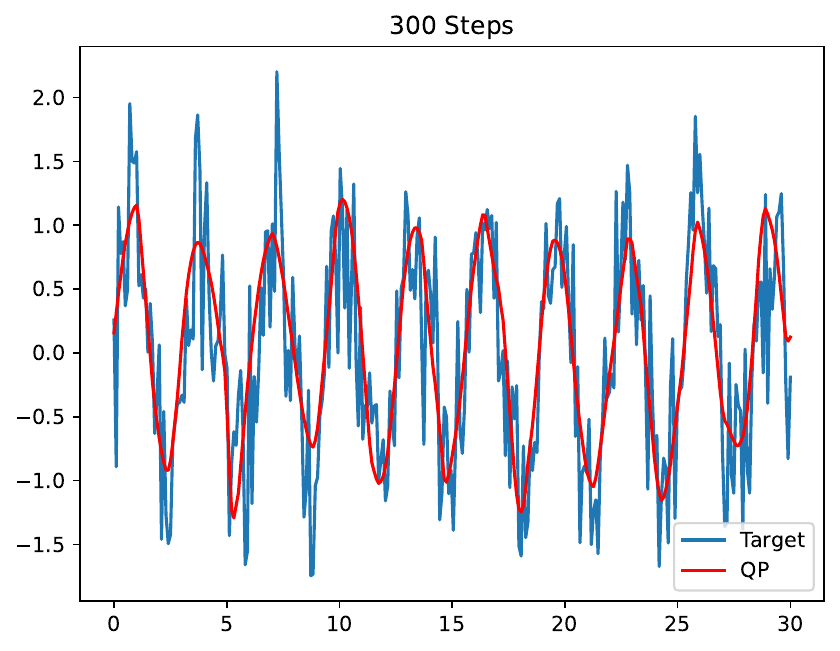}   &\includegraphics[width=0.4\linewidth]{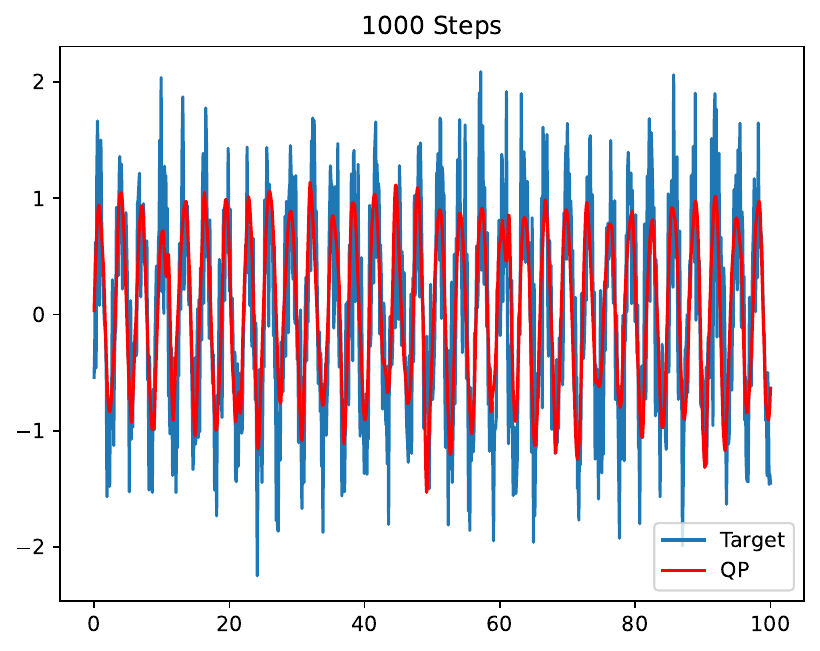}
\end{tabular}
\caption{Comparison of RK4 solver from \textit{torchdiffeq} and our NeuRLP solver for fitting sinusoidal waves with Gaussian noise added at each iteration. Length of the wave and number of steps is 300 (left column) and 1000 (right column). Step size is 0.1. Trained for 100 iterations. The NeuRLP solver has better performance (and efficiency) for longer trajectories.}
\label{fig:solver-benchmark-2}
\end{figure}

We compare NeuRLP with the RK4 solver from \textit{torchdiffeq} on a task of fitting noisy sinusoidal waves of varying lengths.
We compare MSE and time in Table \ref{tab:solver-benchmark} and Figure \ref{fig:solver-benchmark-2}. 

\begin{table}[h]
\caption{Comparing the NeuRLP solver with the RK4 solver with a step size of 0.1 on fitting noisy sinusoidal waves of 300 and 1000 steps. Showing MSE loss and time.}
\begin{center}
\resizebox{0.7\linewidth}{!}{
\begin{tabular}{ ccccc } 
\toprule
Steps&QP (seconds)&RK4 (seconds)& QP Loss & RK4 Loss\\
\midrule
40&1.52&28.06&11.4&29.3\\
100&1.61&64.57&27.9&35.6\\
300&1.76&211.52&52&96.8\\
500&2.12&359.7&128&301\\
1000&3.68&666.69&292&589\\
\bottomrule
\end{tabular}}
\label{tab:solver-benchmark}
\end{center}
\end{table}

\begin{figure}
\centering
  \includegraphics[width=0.4\linewidth]{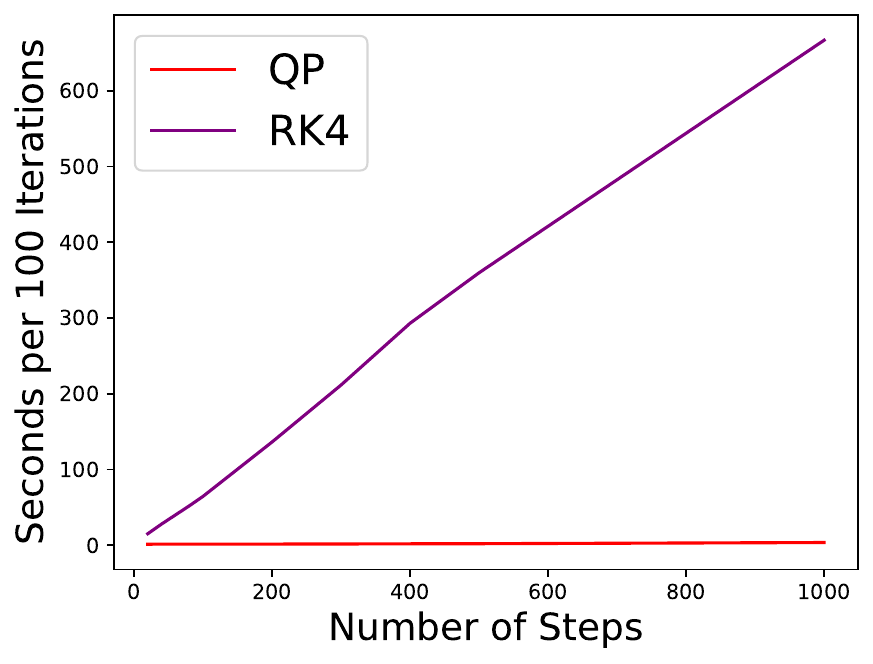} 
  \caption{Number of seconds per 100 iterations for fitting noisy sinusoidal waves. The NeuRLP solver is significantly more efficient over longer times due to its parallelism.}
  \end{figure}

\end{document}